\documentclass[10pt,twocolumn,letterpaper]{article}
\usepackage[pagenumbers]{cvpr}

\usepackage{times}
\usepackage{epsfig}
\usepackage{graphicx}
\usepackage{amsmath,amssymb,amsfonts,amsthm}
\usepackage{bm}
\usepackage{booktabs}
\usepackage{multirow}
\usepackage{xcolor}
\usepackage{array}
\usepackage{tabularx}
\usepackage{enumitem}
\usepackage{xspace}
\usepackage[ruled,vlined]{algorithm2e}
\SetKwInput{KwInput}{Input}
\SetKwInput{KwOutput}{Output}

\newtheorem{proposition}{Proposition}
\newtheorem{theorem}{Theorem}
\newtheorem{lemma}{Lemma}
\newtheorem{corollary}{Corollary}

\newcommand{\method}{RASP\xspace}
\newcommand{\R}{\mathbb{R}}
\newcommand{\E}{\mathbb{E}}
\newcommand{\norm}[1]{\left\|#1\right\|}

\newcommand{\diag}{\mathrm{diag}}
\newcommand{\act}{\mathcal{A}_t}
\newcommand{\inact}{\mathcal{A}_t^c}
\newcommand{\frz}{\mathcal{L}}
\newcommand{\unf}{\mathcal{P}}
\newcommand{\orth}{\mathcal{R}_{\mathrm{dec}}}

\definecolor{cvprblue}{rgb}{0.21,0.49,0.74}
\usepackage[pagebackref,breaklinks,colorlinks,allcolors=cvprblue]{hyperref}

\def\confName{CVPR}
\def\confYear{2026}
\def\paperID{0000}

\title{\method: Read-All, Sparse-Plasticity Continual Learning in a Stationary-Size LoRA}

\author{
Wen-Hsuan Liu\\
University of California, Irvine
\and
Amara Okonkwo\\
University of Washington
\and
Tom Saliencro\\
University of California, Irvine
\and
Rohan Desai\\
University of Washington
\and
Maya Lindqvist\\
University of California, Irvine
}

\begin{document}
\maketitle

\begin{abstract}
Parameter-efficient continual learning on pre-trained Vision Transformers largely relies on a costly premise: preventing catastrophic forgetting requires either spawning a dedicated LoRA branch per session—causing adapter parameters to scale linearly with task horizon—or applying sparse Mixture-of-Experts gating symmetrically to both forward and backward passes, which sacrifices representational rank at test time. We argue that reading consolidated knowledge and writing new task credit are fundamentally asymmetric operations. Because visual features in a shared low-rank space distribute across rank-1 atoms rather than localizing in disjoint task silos, forward inference should read \emph{all} atoms at full rank capacity, whereas backward optimization should restrict plasticity to a sparse subset of input-aligned channels. We introduce \textbf{\method} (\textbf{R}ead-\textbf{A}ll, \textbf{S}parse-\textbf{P}lasticity LoRA), which adapts a single stationary-size adapter across sequential tasks. At each step, a scene-modulated cosine router selects the top-$q$ unlocked atoms to receive gradients, blocks updates on the remaining $r-q$ channels, and penalizes Frobenius Gram coherence between active and non-updated output directions. Across sessions, the router's empirical activation trace and each atom's exact spectral norm form a zero-overhead geometric consolidation index $\omega_j = \sqrt{\tilde{c}_j \tilde{\sigma}_j}$ that drives Cauchy step attenuation and boundary anchor locking. Across CIFAR-100, CUB-200, ImageNet-R, and ImageNet-A, \method matches or surpasses branch-expanding LoRA learners under a strictly constant parameter footprint.
\end{abstract}

\section{Introduction}
\label{sec:intro}

Exemplar-free class-incremental learning (CIL) seeks to absorb a sequence of visual categories without storing rehearsal buffers or querying task identifiers at inference time \cite{zhou2024cilsurvey}. Early continual learners attributed catastrophic forgetting to unconstrained stochastic gradient updates overwriting shared network weights \cite{kirkpatrick2017ewc}. With frozen pre-trained Vision Transformers (ViTs) \cite{dosovitskiy2021vit} now serving as the standard visual backbone, adaptation is predominantly localized inside Low-Rank Adaptation (LoRA) modules \cite{hu2022lora,liu2024dora}, shifting the core design question from full-network regularization to low-rank subspace governance.

To shield earlier sessions from later updates, mainstream LoRA-based continual learners instantiate a fresh adapter branch for every incoming task. InfLoRA initializes each new branch inside the empirical feature null-space of prior sessions \cite{liang2024inflora}; SD-LoRA decouples directional bases from learnable magnitudes across growing branches \cite{wu2025sdlora}; BiLoRA constructs probabilistically orthogonal bilinear subspaces \cite{zhu2025bilora}; PLAN selects low-sensitivity bases via perturbation probing \cite{wang2025plan}; LoRA-P\&M interpolates perturbed task vectors \cite{qiu2025pm}; and SplitLoRA partitions the gradient space into primary and minor subspaces \cite{qiu2026splitlora}. While effective, branch expansion hard-codes two structural liabilities: adapter memory grows linearly without bound as tasks accumulate, and locking each session inside a private branch prevents later tasks from reusing fine-grained visual primitives learned earlier. Conversely, recent intra-adapter routing methods—such as FlyLoRA \cite{zou2025flylora}, NeuroLoRA \cite{yang2026neurolora}, and rank-1 expert pools \cite{fa2026dac}—maintain a shared parameter pool but apply top-$q$ sparsity symmetrically to both forward and backward passes. Zeroing $r-q$ channels during forward evaluation discards $(1 - q/r)$ of the adapter's effective rank capacity on every test sample.

\begin{figure*}[t]
    \centering
    \includegraphics[width=0.98\textwidth]{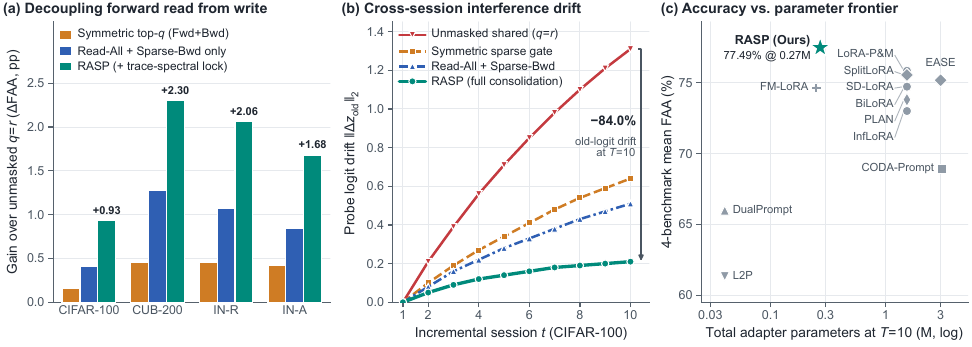}
    \caption{Motivation for \method{}. (a)~Across all four benchmarks, decoupling full-rank forward reading from sparse backward writing consistently outperforms both unmasked shared updates ($q=r$) and symmetric top-$q$ gating, with trace-spectral consolidation providing an additional boost. (b)~On CIFAR-100, unmasked shared training suffers severe old-task logit drift; symmetric gating curbs drift at the expense of forward rank; \method{} suppresses $84.0\%$ of cumulative logit drift while preserving full forward capacity. (c)~At $T=10$ sessions under a unified ViT-B/16 protocol, \method{} occupies the Pareto frontier of final accuracy versus total adapter parameters.}
    \label{fig:motivation}
\end{figure*}

Our key insight is illustrated in Figure~\ref{fig:motivation}. When a single fixed-size LoRA is trained across a task stream, discriminative visual structure does not segregate into mutually exclusive per-task ranks; instead, it organizes as a distributed superposition across all $r$ rank-1 atoms. Consequently, \textbf{forward representation reading and backward gradient writing should not share the same sparsity mask}. During the forward pass, all $r$ atoms should remain active (\emph{Read-All}) so the classifier always operates at full rank capacity and can compose features learned in any prior session. During the backward pass, credit assignment should be restricted to the $q$ unlocked atoms most aligned with the current input (\emph{Sparse-Plasticity}), leaving the remaining $r-q$ channels untouched at that step.

Better still, this backward router generates its own cross-session consolidation signal at zero extra compute. By pairing each atom's cumulative routing trace $\tilde{c}_j$ with its exact rank-1 spectral norm $\tilde{\sigma}_j = \|u_j\|_2 \|v_j\|_2$, we obtain a geometric consolidation index $\omega_j = \sqrt{\tilde{c}_j \tilde{\sigma}_j}$ that identifies atoms that are both frequently recruited across tasks and energetically load-bearing. This single index governs continuous Cauchy step-size attenuation on active unlocked channels and locks a small quota of top-$\omega_j$ atoms into an immutable anchor set at each session boundary.

The main contributions of this work are fourfold:
\begin{itemize}[leftmargin=1.3em]
    \item We formulate stationary-budget LoRA continual learning through \emph{read-write asymmetry}: unmasked rank-$r$ forward superposition preserves representational capacity and cross-task transfer, while routed top-$q$ backward gating localizes per-step plasticity.
    \item We design a scene-modulated, usage-equalized cosine router coupled with a cross-partition Frobenius Gram coherence penalty that steers active output updates away from non-updated and locked directions.
    \item We introduce a zero-overhead \emph{trace-spectral consolidation} mechanism ($\omega_j = \sqrt{\tilde{c}_j \tilde{\sigma}_j}$) with provable false-locking suppression and a three-tier cross-session logit drift bound.
    \item Extensive experiments on CIFAR-100, CUB-200, ImageNet-R, and ImageNet-A under a unified ViT-B/16 protocol show that \method{} matches or outperforms per-task expansion methods at strictly constant adapter size.
\end{itemize}

\section{Related Work}
\label{sec:related}

\paragraph{Pre-trained continual learning.}
Classical continual learning relies on rehearsal buffers, parameter regularization, or network expansion \cite{zhou2024cilsurvey,kirkpatrick2017ewc}. With strong pre-trained ViTs, rehearsal-free methods adapt lightweight modules instead. Prompt-tuning approaches such as L2P, DualPrompt, and CODA-Prompt assemble task-conditioned prompts from learnable pools \cite{wang2022l2p,wang2022dualprompt,smith2023codaprompt}. Adapter and prototype methods—including LAE \cite{gao2023lae}, EASE \cite{zhou2024ease}, and MOS \cite{sun2025mos}—combine lightweight feature adapters with calibrated class prototypes, while analytic learners such as Fly-CL and guided random projections fit closed-form ridge heads over expanded representations \cite{zou2026flycl,li2026grp}. \method{} is orthogonal to classifier-side calibration: it governs how a single low-rank backbone adapter updates and consolidates internal representations across sessions.

\paragraph{Subspace-expanding vs.\ stationary LoRA.}
Most LoRA-based continual learners enforce stability via spatial branch isolation \cite{liang2024inflora,wu2025sdlora,zhu2025bilora,wang2025plan,qiu2025pm,qiu2026splitlora} or inter-branch orthogonal constraints \cite{wang2023olora,singh2026identifiable,liu2026codelora}. Because each session claims a dedicated parameter block, memory scales as $\mathcal{O}(T)$ and early features are locked away from later tasks unless explicitly merged. Recent efforts toward shared low-rank spaces explore learnable routing matrices in C-LoRA \cite{zhang2025clora}, meta-prompted rank selection in FM-LoRA \cite{yu2025fmlora}, paced updates in PaLoRA \cite{li2026palora}, activation-subspace protection in SPACE-LoRA \cite{yoo2026spacelora}, geodesic subspace evolution in Geo-LoRA \cite{feng2026geolora}, spectral balancing in MuLoRA \cite{liu2026mulora}, and post-hoc parameter freezing or task-conditioned transforms \cite{ling2025critical,fu2026oneadapter}. Closer in spirit, E$^2$-LoRA \cite{li2026e2lora} orders task drift via post-session SVD to recycle low-energy ranks, but still requires an extra eigendecomposition pass. In contrast, \method{} operates inside a single stationary-size LoRA and extracts its consolidation signal directly from the router's runtime trace and rank-1 norms.

\paragraph{Mixture-of-Experts and rank-level routing.}
MoE-LoRA architectures route tokens across multiple adapter experts \cite{wu2024mole,qiu2025mingle,lu2026moram}. Moving routing inside a single adapter, FlyLoRA \cite{zou2025flylora} and NeuroLoRA \cite{yang2026neurolora} select top-$q$ ranks via sparse projections or neuromodulated gates, while DAC \cite{fa2026dac} and noise-floor squelching \cite{okonkwo2026squelch} compose sparse rank-1 pools. As recent diagnostics emphasize, routing alone does not eliminate subspace contention within shared adapters \cite{chowdhury2026routing}. Crucially, all prior rank routers apply their sparsity gate symmetrically to the forward pass, truncating inference rank. \method{} keeps forward composition completely dense and confines sparsity, Gram decoupling, and trace-spectral consolidation to the backward update.

\section{Preliminaries}
\label{sec:prelim}

We study rehearsal-free class-incremental learning over $T$ sequential sessions $\{\mathcal{T}_1, \dots, \mathcal{T}_T\}$. Session $t$ provides training dataset $\mathcal{D}_t = \{(x_i, y_i)\}_{i=1}^{n_t}$ with label set $\mathcal{Y}_t$, where $\mathcal{Y}_t \cap \mathcal{Y}_{s} = \emptyset$ for $t \neq s$. No historical exemplars are retained, and test-time evaluation requires classifying across all seen categories $\mathcal{Y}_{1:t} = \bigcup_{s=1}^t \mathcal{Y}_s$ without task identity.

Let $W_0 \in \R^{d \times d}$ denote a frozen pre-trained projection layer (key or value matrix in ViT-B/16). Instead of allocating task-indexed adapters $\{B_t, A_t\}_{t=1}^T$, \method{} maintains a single stationary-size low-rank module parameterized by input probe matrix $V \in \R^{d \times r}$ and output synthesis matrix $U \in \R^{d \times r}$ ($r \ll d$). Writing $V = [v_1, \dots, v_r]$ and $U = [u_1, \dots, u_r]$, the weight update decomposes into $r$ rank-1 atomic channels:
\begin{equation}
\label{eq:atomic_lora}
\Delta W = \frac{1}{r} U V^\top = \frac{1}{r}\sum_{j=1}^{r} u_j v_j^\top,
\qquad
h = W_0 x + \frac{1}{r}\sum_{j=1}^{r} u_j (v_j^\top x).
\end{equation}
Here $v_j^\top x \in \R$ measures how strongly token $x \in \R^d$ excites the $j$-th input probe, and $u_j \in \R^d$ specifies the output synthesis direction added to the residual stream. Throughout training on session $t$, the input probes $V$ are held fixed while only selected columns of $U$ and a lightweight scene gate are updated.

\section{\method{} Framework}
\label{sec:method}

\begin{figure*}[t]
    \centering
    \includegraphics[width=\textwidth]{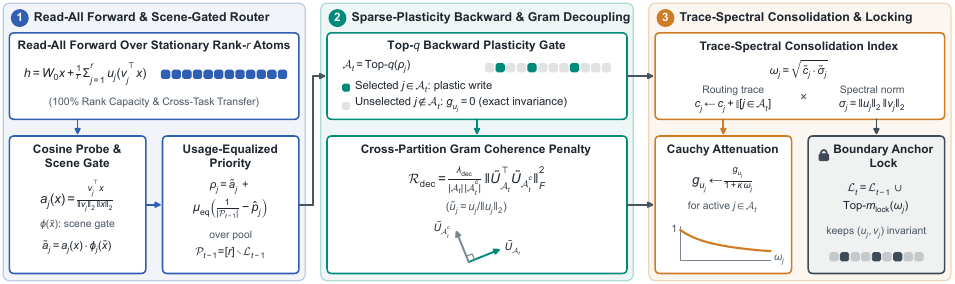}
    \caption{Overview of the \method{} pipeline. \textbf{Stage 1 (Left):} Every input batch traverses all $r$ rank-1 atoms in the forward pass without truncation; a cosine probe modulated by an image-level scene gate $\phi(\bar{x})$ and a usage-equalizing prior scores the unlocked pool $\unf_{t-1}$. \textbf{Stage 2 (Middle):} Only the top-$q$ unlocked channels $\act$ receive backward gradients, while a Frobenius Gram coherence penalty $\orth$ pushes active synthesis columns orthogonal to non-updated and locked columns. \textbf{Stage 3 (Right):} Cumulative routing trace $\tilde{c}_j$ and rank-1 spectral norm $\tilde{\sigma}_j$ form the consolidation index $\omega_j = \sqrt{\tilde{c}_j \tilde{\sigma}_j}$, driving continuous Cauchy step attenuation and boundary anchor locking.}
    \label{fig:architecture}
\end{figure*}

\method{} resolves the stability--plasticity dilemma inside a stationary rank-$r$ adapter by coupling two mechanisms into a closed loop (Figure~\ref{fig:architecture}): (i)~intra-session \emph{Read-All, Sparse-Plasticity} updates with cross-partition Gram decoupling, and (ii)~inter-session \emph{Trace-Spectral Consolidation} that converts routing history into graded step attenuation and permanent anchor locking.

\subsection{Read-All Forward \& Scene-Gated Router}
\label{sec:router}

In the forward pass, all $r$ atomic channels contribute via Equation~\eqref{eq:atomic_lora}. No channel is dropped during feature extraction or test-time prediction, ensuring that visual primitives encoded in earlier sessions remain continuously accessible for cross-task composition.

To decide which channels should absorb backward credit on the current minibatch, we first compute a scale-invariant directional cosine probe for each channel $j \in \{1, \dots, r\}$:
\begin{equation}
\label{eq:cosine_probe}
a_j(x) = \frac{|v_j^\top x|}{\norm{v_j}_2 \norm{x}_2 + \epsilon}.
\end{equation}
Dividing by $\norm{v_j}_2 \norm{x}_2$ prevents probe norm variations or token magnitude spikes from biasing channel selection.

Because Equation~\eqref{eq:cosine_probe} evaluates tokens locally, two patches from semantically different categories (e.g., background foliage vs.\ bird plumage) could trigger identical probes. To inject global scene context at negligible cost, we form the RMS-normalized token summary $\bar{x} = \mathrm{RMSNorm}\!\big(\frac{1}{N}\sum_{n=1}^N x_n\big) \in \R^d$ and pass it through a lightweight bottleneck modulator ($d_g = 64$, $\sim 0.10$M fixed parameters across the network):
\begin{equation}
\label{eq:scene_gate}
\phi(\bar{x}) = \mathbf{1} + \tanh\!\big(W_2\,\mathrm{SiLU}(W_1 \bar{x})\big) \odot \xi,
\end{equation}
where $W_1 \in \R^{d_g \times d}$, $W_2 \in \R^{r \times d_g}$, and $\xi \in \R^r$ is initialized to zero so $\phi(\bar{x}) = \mathbf{1}$ at the beginning of training. The scene-modulated affinity is $\tilde{a}_j(x) = a_j(x) \cdot \phi_j(\bar{x})$.

Finally, to prevent early winner-take-all routing collapse among unlocked channels $\unf_{t-1} = \{1, \dots, r\} \setminus \frz_{t-1}$, we maintain an exponential moving average $\hat{p}_j$ of each channel's selection probability and add a centered usage-equalizing prior:
\begin{equation}
\label{eq:routing_score}
\rho_j = \tilde{a}_j(x) + \mu_{\mathrm{eq}}\!\left(\frac{1}{|\unf_{t-1}|} - \hat{p}_j\right),
\qquad
j \in \unf_{t-1},
\end{equation}
where $\mu_{\mathrm{eq}} > 0$ boosts under-explored channels during early epochs while letting data-driven affinity dominate once representations specialize.

\subsection{Sparse-Plasticity Backward \& Gram Decoupling}
\label{sec:backward}

Given routing scores $\{\rho_j : j \in \unf_{t-1}\}$ aggregated over the minibatch, \method{} selects the top-$q_t$ unlocked channels ($q_t = \min(q, |\unf_{t-1}|)$) as the active write set $\act$ and applies a hard binary stop-gradient mask to all other synthesis columns:
\begin{equation}
\label{eq:sparse_grad}
\act = \mathrm{Top}\text{-}q_t\!\big(\{\rho_j : j \in \unf_{t-1}\}\big),
\qquad
g_{u_j} = 0 \quad \forall j \notin \act,
\end{equation}
where $g_{u_j} = \nabla_{u_j}\mathcal{L}_{\mathrm{ce}}$ is the backpropagated cross-entropy gradient. The plasticity ratio $q/r$ directly governs how much of the shared subspace a single minibatch may perturb: $r-q_t$ channels experience exact zero parameter drift at every step, while all $r$ channels participate in the forward prediction, unlike the symmetric mask of FlyLoRA \cite{zou2025flylora}.

Masking gradients on $\inact = \{1, \dots, r\} \setminus \act$ freezes their coordinates for the current step, but does not prevent the actively updated columns $\{u_i : i \in \act\}$ from rotating into the subspace already spanned by $\inact$. To enforce geometric separation between plastic updates and protected knowledge, we penalize the Frobenius norm of the cross-partition Gram matrix between unit-normalized active columns $\tilde{U}_{\act}$ and non-active columns $\tilde{U}_{\inact}$ ($\tilde{u}_j = u_j / \norm{u_j}_2$):
\begin{equation}
\label{eq:gram_dec}
\orth = \frac{\lambda_{\mathrm{dec}}}{|\act|\,|\inact|}\,\norm{\tilde{U}_{\act}^\top \tilde{U}_{\inact}}_F^2.
\end{equation}
Unlike all-pairs orthogonalization, Equation~\eqref{eq:gram_dec} leaves correlations \emph{within} $\act$ unconstrained so co-selected channels can cooperatively fit the current task, while repelling their updates from inactive and locked anchors.

\subsection{Trace-Spectral Consolidation \& Locking}
\label{sec:consolidation}

Stepwise masking protects channels outside $\act$, yet an unlocked channel crucial to Session~$1$ may still be routed into $\act$ during Session~$4$ and overwritten. Rather than introducing a separate Fisher pass \cite{kirkpatrick2017ewc} or post-session SVD \cite{li2026e2lora}, \method{} harvests a consolidation signal directly from two quantities intrinsic to the routing loop.

First, every top-$q_t$ selection emits a binary indicator $\mathbb{I}[j \in \act]$, which we accumulate into a cumulative routing trace across all training steps:
\begin{equation}
\label{eq:trace}
c_j \leftarrow c_j + \mathbb{I}[j \in \act], \qquad j = 1, \dots, r.
\end{equation}
Second, while $c_j$ records how consistently channel $j$ is recruited, it does not measure how strongly atom $u_j v_j^\top$ modifies the frozen backbone. By Proposition~\ref{prop:spectral}, the exact operator (and nuclear) norm of the $j$-th rank-1 atom is
\begin{equation}
\label{eq:spec_norm}
\sigma_j = \norm{u_j v_j^\top}_2 = \norm{u_j}_2 \norm{v_j}_2.
\end{equation}
We normalize both quantities to $[0,1]$ via $\tilde{c}_j = c_j / \max_k c_k$ and $\tilde{\sigma}_j = \sigma_j / \max_k \sigma_k$, and define their geometric mean as the \textbf{trace-spectral consolidation index}:
\begin{equation}
\label{eq:omega}
\omega_j = \sqrt{\tilde{c}_j \cdot \tilde{\sigma}_j} \in [0, 1].
\end{equation}
A channel attains a high $\omega_j$ if and only if it is both frequently selected across tasks and spectrally energetic. Noisy tie-breaker channels ($\tilde{c}_j \gg 0, \tilde{\sigma}_j \approx 0$) and stale one-task spikes ($\tilde{c}_j \approx 0, \tilde{\sigma}_j \gg 0$) are automatically suppressed.

\paragraph{Intra-session Cauchy step attenuation.}
For unlocked channels selected into $\act$, gradients are rescaled inversely by their consolidation index consolidated at the end of session $t-1$:
\begin{equation}
\label{eq:cauchy}
g_{u_j} \leftarrow \frac{g_{u_j}}{1 + \kappa\,\omega_j^{(t-1)}}, \qquad j \in \act,
\end{equation}
where $\kappa > 0$ is the attenuation gain. Because $\omega_j$ already incorporates the square-root geometric compression of Equation~\eqref{eq:omega}, the Cauchy multiplier $(1 + \kappa\,\omega_j)^{-1}$ smoothly decelerates updates on shared high-utility atoms without numerically zeroing their plasticity.

\paragraph{Inter-session boundary anchor locking.}
At the end of session $t$, we refresh $\sigma_j$ and $\omega_j^{(t)}$. To prevent a single high-gradient step in future tasks from corrupting the most critical consolidated atoms, we permanently lock the top $m_{\mathrm{lock}}$ unlocked channels into the anchor set $\frz_t$:
\begin{equation}
\label{eq:lock}
\frz_t = \frz_{t-1} \cup \mathrm{Top}\text{-}m_{\mathrm{lock}}\!\big(\{\omega_j^{(t)} : j \in \unf_{t-1}\}\big),
\end{equation}
where $m_{\mathrm{lock}} = \mathrm{clip}(\lfloor \eta_{\mathrm{lock}} |\unf_{t-1}| \rfloor, 1, 3)$ with $\eta_{\mathrm{lock}} = 0.05$, and locking halts once $|\unf_{t-1}| < q + 2$ to prevent plasticity starvation. Crucially, at the start of session $t+1$, only the \emph{unlocked} input probes $\{v_j : j \in \unf_t\}$ are re-initialized in the empirical feature null-space of prior tasks \cite{liang2024inflora}, whereas locked pairs $\{(u_j, v_j) : j \in \frz_t\}$ remain completely invariant in both factors. Algorithm~\ref{alg:rasp} summarizes the complete procedure.

\begin{algorithm}[t]
\small
\caption{\method{} Training Procedure at Session $t$}
\label{alg:rasp}
\KwInput{Shared adapter $(U, V)$, scene gate $\phi$, locked anchor set $\frz_{t-1}$, trace $c$, session data $\mathcal{D}_t$, budget $q$, gain $\kappa$}
\KwOutput{Updated $(U, V, \frz_t, c, \omega^{(t)})$}

\tcp{Null-space probe alignment on unlocked pool only}
$\unf_{t-1} \leftarrow \{1, \dots, r\} \setminus \frz_{t-1}$ \;
Project $\{v_j : j \in \unf_{t-1}\}$ onto null-space of prior feature covariance; keep $\{v_j : j \in \frz_{t-1}\}$ intact \;

\ForEach{minibatch $(X, Y) \subset \mathcal{D}_t$}{
  Compute Read-All forward output $h$ over all $r$ atoms via Eq.~\eqref{eq:atomic_lora} \;
  Evaluate cosine probes $a_j(x)$, scene gate $\phi(\bar{x})$, and scores $\rho_j$ on $\unf_{t-1}$ via Eqs.~\eqref{eq:cosine_probe}--\eqref{eq:routing_score} \;
  Select $\act \leftarrow \mathrm{Top}\text{-}q_t(\{\rho_j : j \in \unf_{t-1}\})$ and mask $g_{u_j} = 0$ for $j \notin \act$ \;
  Compute gradient of $\mathcal{L}_{\mathrm{ce}} + \orth$ (Eq.~\eqref{eq:gram_dec}) and apply Cauchy attenuation $g_{u_j} \leftarrow (1 + \kappa\,\omega_j^{(t-1)})^{-1} g_{u_j}$ \;
  Update active synthesis atoms $\{u_j : j \in \act\}$ and scene gate; accumulate $c_j \leftarrow c_j + \mathbb{I}[j \in \act]$ \;
}
Update spectral norms $\sigma_j = \norm{u_j}_2 \norm{v_j}_2$ and consolidation indices $\omega_j^{(t)} = \sqrt{\tilde{c}_j \tilde{\sigma}_j}$ \;
\If{$|\unf_{t-1}| \ge q + 2$}{
  $\frz_t \leftarrow \frz_{t-1} \cup \mathrm{Top}\text{-}m_{\mathrm{lock}}(\{\omega_j^{(t)} : j \in \unf_{t-1}\})$ \;
}
\Return{$(U, V, \frz_t, c, \omega^{(t)})$}
\end{algorithm}

\section{Theoretical Analysis}
\label{sec:theory}

We now formalize why decoupling forward reading from backward writing preserves representational rank while bounding cross-session interference. Full proofs are provided in Appendix~\ref{sec:supp-proofs}.

\begin{proposition}[Read-write capacity--interference decoupling]
\label{prop:decoupling}
For any active write subset $\act \subset \{1, \dots, r\}$ of cardinality $q$, let $\Delta W_{\mathrm{sym}} = \frac{1}{r}\sum_{j \in \act} u_j v_j^\top$ denote the symmetrically gated adapter. For any input $x \in \R^d$, the forward truncation error incurred by symmetric gating satisfies
\begin{equation}
\label{eq:trunc_gap}
\norm{(\Delta W - \Delta W_{\mathrm{sym}})x}_2
\le
\frac{\norm{x}_2}{r}\sum_{j \notin \act} \sigma_j\,a_j(x),
\end{equation}
whereas \method{} incurs zero forward truncation error ($\Delta W_{\mathrm{fwd}} = \Delta W$). Simultaneously, under isotropic per-channel gradient variance $\varsigma^2$, \method{}'s sparse-plasticity mask reduces expected backward update energy from $r\varsigma^2$ to $q\varsigma^2$.
\end{proposition}

Proposition~\ref{prop:decoupling} exposes the structural flaw of symmetric rank routing \cite{zou2025flylora}: zeroing $r-q$ channels in the forward pass discards a feature residual proportional to the spectral norms $\sigma_j$ of unselected atoms, whereas \method{} eliminates $(1 - q/r)$ of backward perturbation energy without sacrificing forward rank.

\begin{theorem}[Three-tier session drift bound]
\label{thm:drift}
Consider session $t$ with locked anchor set $\frz_{t-1}$ and unlocked pool $\unf_{t-1}$. Let $S_j \subset \{1, \dots, M_t\}$ denote the minibatch steps at which channel $j \in \unf_{t-1}$ is routed into $\act$ with step sizes $\{\eta_s\}$. Because locked pairs $\{(u_j, v_j) : j \in \frz_{t-1}\}$ and unlocked probes $\{v_j : j \in \unf_{t-1}\}$ remain fixed during session $t$, the functional output shift $\Delta h_t(x) = \Delta W^{(t)}x - \Delta W^{(t-1)}x$ on any probe input $x$ decomposes into three tiers:
\begin{equation}
\label{eq:three_tier}
\norm{\Delta h_t(x)}_2
\le
\frac{\norm{x}_2}{r}\!\sum_{\substack{j \in \unf_{t-1} \\ S_j \neq \emptyset}}
\frac{\norm{v_j}_2\,a_j(x)}{1 + \kappa\,\omega_j^{(t-1)}}
\sum_{s \in S_j} \eta_s \norm{g_{u_j}^{(s)}}_2,
\end{equation}
with exact zero drift contributed by locked anchors ($j \in \frz_{t-1}$) and unrouted unlocked channels ($j \in \unf_{t-1}, S_j = \emptyset$).
\end{theorem}

Theorem~\ref{thm:drift} guarantees that locked anchors ($\frz_{t-1}$) and unrouted unlocked atoms ($S_j = \emptyset$) contribute \emph{exact zero} functional drift, while drift on revisited channels is damped by $(1 + \kappa\,\omega_j^{(t-1)})^{-1}$.

\begin{proposition}[Spectral identity and conjunctive outlier filtering]
\label{prop:spectral}
Each rank-1 atom $M_j = u_j v_j^\top$ satisfies $\norm{M_j}_2 = \norm{M_j}_* = \norm{u_j}_2 \norm{v_j}_2 = \sigma_j$. Moreover, for any locking threshold $\tau \in (0, 1)$, the geometric consolidation condition $\omega_j = \sqrt{\tilde{c}_j \tilde{\sigma}_j} > \tau$ enforces the conjunctive lower bound $\min(\tilde{c}_j, \tilde{\sigma}_j) > \tau^2$, whereas the arithmetic mean $\frac{1}{2}(\tilde{c}_j + \tilde{\sigma}_j) > \tau$ admits single-cue false positives with $\min(\tilde{c}_j, \tilde{\sigma}_j) = 0$ whenever $\tau < 0.5$.
\end{proposition}

\section{Experiments}
\label{sec:experiments}

\begin{table*}[t]
    \centering
    \footnotesize
    \setlength{\tabcolsep}{2.2pt}
    \caption{Comparison with state-of-the-art rehearsal-free CIL methods across four 10-session benchmarks on ViT-B/16-IN21K. ``$\Delta$Params'' denotes additional adapter parameters allocated per session, and ``Total@$T_{10}$'' is the cumulative adapter overhead after 10 sessions ($^\dagger$: $0.25$M for $r=50$, $0.30$M for $r=64$, $0.35$M for $r=80$, including the fixed $0.10$M scene gate). All cells are evaluated under a unified 3-seed protocol. \textbf{Bold} and \underline{underline} indicate best and second-best among LoRA-based methods.}
    \label{tab:main}
    \begin{tabular*}{\textwidth}{@{\extracolsep{\fill}}llcccccccccc@{}}
        \toprule
        & & & & \multicolumn{2}{c}{CIFAR-100} & \multicolumn{2}{c}{CUB-200} & \multicolumn{2}{c}{ImageNet-R} & \multicolumn{2}{c}{ImageNet-A} \\
        \cmidrule(lr){5-6}\cmidrule(lr){7-8}\cmidrule(lr){9-10}\cmidrule(lr){11-12}
        Method & Venue & $\Delta$Params & Total@$T_{10}$ & FAA & CAA & FAA & CAA & FAA & CAA & FAA & CAA \\
        \midrule
        \multicolumn{12}{@{}l}{\textit{Prompt-tuning families}} \\
        L2P \cite{wang2022l2p} & CVPR'22 & +0 & 0.04M & 82.79 & 88.45 & 62.28 & 73.89 & 66.54 & 72.81 & 33.84 & 45.91 \\
        DualPrompt \cite{wang2022dualprompt} & ECCV'22 & +0 & 0.04M & 85.61 & 90.36 & 66.08 & 77.96 & 68.47 & 72.64 & 33.79 & 46.83 \\
        CODA-Prompt \cite{smith2023codaprompt} & CVPR'23 & +0.31M & 3.07M & 86.65 & 91.12 & 74.83 & 83.06 & 71.52 & 75.86 & 42.79 & 53.61 \\
        \midrule
        \multicolumn{12}{@{}l}{\textit{Adapter \& prototype-surgery families}} \\
        LAE \cite{gao2023lae} & ICCV'23 & +0.19M & 1.90M & 85.32 & 90.41 & 75.64 & 83.72 & 72.68 & 77.95 & 45.12 & 55.84 \\
        EASE \cite{zhou2024ease} & CVPR'24 & +0.30M & 3.00M & 86.58 & 91.71 & 79.94 & 86.69 & 77.79 & 83.91 & 56.45 & 66.61 \\
        MOS \cite{sun2025mos} & AAAI'25 & +0.30M & 3.00M & 89.29 & 93.34 & 89.86 & 92.82 & 77.96 & 83.01 & 55.26 & 65.82 \\
        \midrule
        \multicolumn{12}{@{}l}{\textit{LoRA with per-session branch expansion}} \\
        InfLoRA \cite{liang2024inflora} & CVPR'24 & +0.15M & 1.53M & 85.24 & 90.61 & \underline{79.92} & \textbf{88.21} & 77.63 & 83.48 & 49.26 & 60.98 \\
        SD-LoRA \cite{wu2025sdlora} & ICLR'25 & +0.15M & 1.53M & 88.05 & 92.58 & 77.54 & 85.63 & 77.38 & 82.09 & 56.01 & 65.02 \\
        BiLoRA \cite{zhu2025bilora} & CVPR'25 & +0.15M & 1.53M & 87.51 & 92.47 & 78.12 & 85.94 & 77.98 & 81.56 & 55.04 & 64.48 \\
        PLAN \cite{wang2025plan} & ICCV'25 & +0.15M & 1.53M & 87.59 & 92.25 & 77.65 & 85.42 & 75.31 & 80.46 & 54.68 & 63.92 \\
        LoRA-P\&M \cite{qiu2025pm} & NeurIPS'25 & +0.15M & 1.53M & 88.49 & 92.92 & 78.34 & 83.45 & \underline{79.92} & \underline{85.24} & 56.63 & 65.41 \\
        SplitLoRA \cite{qiu2026splitlora} & ICLR'26 & +0.15M & 1.53M & \underline{90.24} & \textbf{93.66} & 72.88 & 83.18 & \textbf{80.96} & \textbf{85.79} & \underline{57.98} & \underline{68.97} \\
        \midrule
        \multicolumn{12}{@{}l}{\textit{LoRA with stationary parameter budget}} \\
        C-LoRA \cite{zhang2025clora} & arXiv'25 & +0 & 0.24M & 87.84 & 92.15 & 76.92 & 84.88 & 76.85 & 81.74 & 53.92 & 63.45 \\
        FM-LoRA \cite{yu2025fmlora} & CVPRW'25 & +0 & 0.25M & 88.41 & 92.69 & 77.84 & 85.96 & 77.91 & 82.63 & 54.45 & 64.18 \\
        \textbf{\method{} (Ours)} & -- & \textbf{+0} & \textbf{0.25M$^\dagger$} & \textbf{90.41}$\pm$.04 & \underline{93.36}$\pm$.08 & \textbf{81.36}$\pm$.22 & \underline{85.68}$\pm$.34 & 79.68$\pm$.14 & 83.52$\pm$.15 & \textbf{58.52}$\pm$.09 & \textbf{70.26}$\pm$.13 \\
        \bottomrule
    \end{tabular*}
\end{table*}

\subsection{Experimental Setup}
\label{sec:setup}

\paragraph{Benchmarks and evaluation protocol.}
We evaluate \method{} on four standard CIL benchmarks spanning natural objects, fine-grained taxonomy, artistic domain shifts, and natural adversarial examples: \textbf{CIFAR-100} \cite{krizhevsky2009cifar}, \textbf{CUB-200-2011} \cite{wah2011cub}, \textbf{ImageNet-R} \cite{hendrycks2021imagenetr}, and \textbf{ImageNet-A} \cite{hendrycks2021imageneta}. Following the standard rehearsal-free protocol \cite{liang2024inflora,sun2025mos}, each dataset is partitioned into $T=10$ sequential sessions with equal class splits ($10$ classes/session on CIFAR-100; $20$ classes/session on CUB-200, ImageNet-R, and ImageNet-A), plus a $T=20$ long-horizon split on ImageNet-R ($10$ classes/session). We report \textbf{FAA} (Final Average Accuracy $A_T$ after the last session), \textbf{CAA} (Cumulative Average Accuracy $\frac{1}{T}\sum_{t=1}^T A_t$), and \textbf{Forgetting} ($F_T$, mean accuracy drop across prior sessions).

\paragraph{Implementation details.}
All methods use a ViT-B/16 backbone pre-trained on ImageNet-21K \cite{dosovitskiy2021vit,ridnik2021imagenet21k} with LoRA modules inserted into the key and value projections of all $12$ blocks. We optimize $U$ and the scene gate using Adam ($\beta_1=0.9, \beta_2=0.999$) with cosine learning-rate decay and batch size $64$. Table~\ref{tab:hparams} summarizes the benchmark-specific operating parameters $(r, q, \text{lr}, \text{epochs}, \kappa)$. Across all benchmarks, we fix scene-gate width $d_g = 64$, usage-prior weight $\mu_{\mathrm{eq}} = 0.1$, Gram decoupling weight $\lambda_{\mathrm{dec}} = 0.3$, and boundary locking fraction $\eta_{\mathrm{lock}} = 0.05$ clipped to $[1, 3]$. The stationary parameter budget of \method{} is a single LoRA ($0.15$M for $r=50$, $0.20$M for $r=64$, $0.25$M for $r=80$) plus the $0.10$M scene gate. All results in our comparison tables are averaged over three random seeds.

\begin{table}[t]
    \centering
    \footnotesize
    \setlength{\tabcolsep}{4.5pt}
    \caption{Benchmark-specific operating configurations for \method{}. Total rank $r$, active plasticity quota $q$, base learning rate, training epochs per session, and Cauchy attenuation gain $\kappa$.}
    \label{tab:hparams}
    \begin{tabular}{lccccc}
        \toprule
        Benchmark & $r$ & $q$ ($q/r$) & lr & Epochs & $\kappa$ \\
        \midrule
        CIFAR-100 & 50 & 6 (12.0\%) & $5\mathrm{e}{-4}$ & 20 & 0.05 \\
        CUB-200 & 50 & 15 (30.0\%) & $5\mathrm{e}{-4}$ & 20 & 0.05 \\
        ImageNet-R & 64 & 18 (28.1\%) & $1\mathrm{e}{-3}$ & 30 & 0.08 \\
        ImageNet-A & 80 & 50 (62.5\%) & $5\mathrm{e}{-4}$ & 20 & 0.05 \\
        \bottomrule
    \end{tabular}
\end{table}

\subsection{Main Benchmark Comparison}
\label{sec:main_results}

\begin{figure*}[t]
    \centering
    \includegraphics[width=\textwidth]{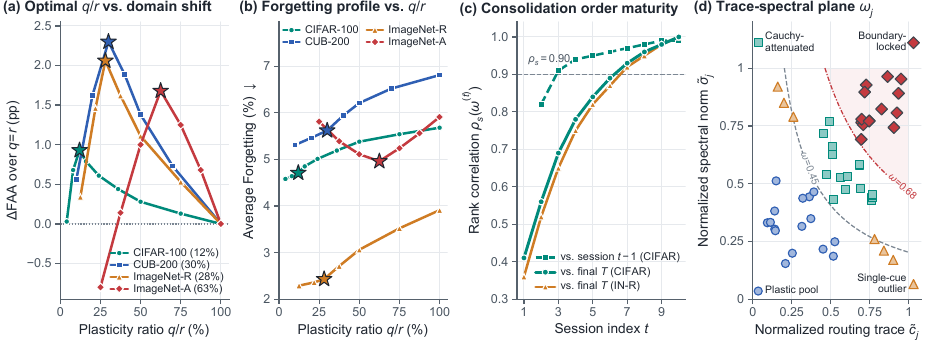}
    \caption{Mechanistic analysis of \method{}. (a, b)~Effect of plasticity ratio $q/r$ on $\Delta$FAA (relative to $q=r$) and Average Forgetting across four benchmarks; stars mark the operating configurations in Table~\ref{tab:hparams}. The optimal $q/r$ shifts rightward as domain divergence from ImageNet-21K increases. (c)~Spearman rank correlation $\rho_s$ of consolidation indices $\omega^{(t)}$ against session $t-1$ and final session $T$, showing rapid stabilization after session 3. (d)~In the 2D trace-spectral plane $(\tilde{c}_j, \tilde{\sigma}_j)$, hyperbolic contours of $\omega_j = \sqrt{\tilde{c}_j \tilde{\sigma}_j}$ isolate true high-trace, high-energy anchors (crimson diamonds) while rejecting single-cue outliers (amber triangles).}
    \label{fig:mechanism}
\end{figure*}

Table~\ref{tab:main} reports the complete comparison across all four benchmarks, with every baseline evaluated across all eight accuracy columns. Three findings stand out:

\paragraph{Outperforming branch expansion at stationary memory.}
On three of the four benchmarks, \method{} achieves the highest final accuracy (FAA) among all LoRA methods without adding parameters across sessions. On CIFAR-100, \method{} reaches $90.41\%$ FAA with $0.25$M total parameters, surpassing SplitLoRA ($90.24\%$, $1.53$M parameters). On ImageNet-A, \method{} reaches $58.52\%$ FAA and $70.26\%$ CAA, outperforming both SplitLoRA ($57.98\%$ / $68.97\%$) and LoRA-P\&M ($56.63\%$ / $65.41\%$). Against stationary-budget peers (C-LoRA and FM-LoRA), \method{} gains $+2.00$ pp on CIFAR-100, $+3.52$ pp on CUB-200, $+1.77$ pp on ImageNet-R, and $+4.07$ pp on ImageNet-A.

\paragraph{Cross-session feature reuse on fine-grained streams.}
The advantage of Read-All forward superposition is clearest on CUB-200. Because bird species share local anatomical attributes (beak geometry, wing barring, plumage texture), isolating each session inside a private branch (SplitLoRA) starves later tasks of reusable primitives, dropping FAA to $72.88\%$. By reading all $r$ shared atoms in the forward pass while protecting established directions in the backward pass, \method{} reaches $81.36\%$ FAA—beating SplitLoRA by $+8.48$ pp, LoRA-P\&M by $+3.02$ pp, and InfLoRA by $+1.44$ pp.

\paragraph{Behavior under multi-style rendition shifts.}
ImageNet-R is the only benchmark where branch expansion retains a narrow lead ($80.96\%$ for SplitLoRA and $79.92\%$ for LoRA-P\&M vs.\ $79.68\%$ for \method{}), while \method{} beats InfLoRA ($77.63\%$), BiLoRA ($77.98\%$), SD-LoRA ($77.38\%$), and PLAN ($75.31\%$). Artistic renditions require a wider plasticity quota ($q/r = 28.1\%$ vs.\ $12.0\%$ on CIFAR-100), leaving fewer channels unrouted per step ($71.9\%$ vs.\ $88.0\%$). However, on the $T=20$ ImageNet-R stream (Table~\ref{tab:long_horizon}), expansion methods double their overhead to $3.06$M ($10.2\times$ \method{}'s fixed $0.30$M), while the gap to SplitLoRA shrinks from $1.28$ pp to $0.63$ pp ($78.79\%$ vs.\ $78.16\%$).

\begin{table}[t]
    \centering
    \footnotesize
    \setlength{\tabcolsep}{3.6pt}
    \caption{Long-horizon scalability on ImageNet-R ($T=20$ sessions, 10 classes/session). ``Total@$T_{20}$'' is the total adapter parameter count after session 20, and $F_{20}$ denotes final forgetting.}
    \label{tab:long_horizon}
    \begin{tabular}{llcccc}
        \toprule
        Method & Venue & Total@$T_{20}$ & FAA & CAA & $F_{20}$ $\downarrow$ \\
        \midrule
        InfLoRA \cite{liang2024inflora} & CVPR'24 & 3.06M & 71.08 & 77.34 & 6.42 \\
        PLAN \cite{wang2025plan} & ICCV'25 & 3.06M & 71.15 & 77.98 & 6.18 \\
        LoRA-P\&M \cite{qiu2025pm} & NeurIPS'25 & 3.06M & 76.94 & 82.85 & 4.12 \\
        SplitLoRA \cite{qiu2026splitlora} & ICLR'26 & 3.06M & \textbf{78.79} & \textbf{84.52} & 3.45 \\
        FM-LoRA \cite{yu2025fmlora} & CVPRW'25 & \textbf{0.25M} & 74.62 & 80.41 & 4.86 \\
        \textbf{\method{} (Ours)} & -- & \textbf{0.30M} & \underline{78.16} & \underline{83.64} & \textbf{3.18} \\
        \bottomrule
    \end{tabular}
\end{table}

\subsection{Ablation and Mechanistic Analysis}
\label{sec:ablation}

\paragraph{Dissecting read-write asymmetry and router design.}
Table~\ref{tab:ablation_arch} evaluates each component of Stage~1 on CIFAR-100 and CUB-200. Starting from an unmasked shared LoRA ($q=r$, Row~1), applying top-$q$ gating symmetrically to both passes (Row~2, as in FlyLoRA \cite{zou2025flylora}) reduces CIFAR-100 forgetting from $5.68\%$ to $5.01\%$ but yields only $+0.16$ pp FAA because forward truncation discards $(1-q/r)$ of inference rank. Switching to \emph{Read-All forward + Sparse-Plasticity backward} (Row~3) immediately lifts FAA to $89.89\%$ on CIFAR-100 and $80.34\%$ on CUB-200. Adding the scene gate $\phi(\bar{x})$ alone (Row~4) or Gram decoupling $\orth$ alone (Row~5) brings modest gains, whereas combining both (Row~6) jumps to $90.41\%$ on CIFAR-100 and $81.36\%$ on CUB-200: scene modulation selects contextually coherent channel subsets, and Gram decoupling prevents those active columns from drifting into inactive subspaces. Row~7 replaces null-space probe alignment on unlocked channels with Gaussian initialization; accuracy drops by $0.57$ pp on CIFAR-100 ($89.84\%$) but still solidly beats unmasked shared training ($89.48\%$).

\begin{table}[t]
    \centering
    \footnotesize
    \setlength{\tabcolsep}{2.8pt}
    \caption{Stage-1 architectural ablation on CIFAR-100 and CUB-200. Every row includes trace-spectral consolidation in Stage~2; Row~7 replaces null-space probe initialization on unlocked channels $\unf_{t-1}$ with Kaiming normal initialization.}
    \label{tab:ablation_arch}
    \begin{tabular}{lcccc}
        \toprule
        & \multicolumn{2}{c}{CIFAR-100} & \multicolumn{2}{c}{CUB-200} \\
        \cmidrule(lr){2-3}\cmidrule(lr){4-5}
        Configuration & FAA & Forget $\downarrow$ & FAA & Forget $\downarrow$ \\
        \midrule
        1. Unmasked shared LoRA ($q=r$) & 89.48 & 5.68 & 79.06 & 6.81 \\
        2. Symmetric sparse (Fwd \& Bwd) & 89.64 & 5.01 & 79.52 & 6.24 \\
        3. Read-All Fwd + Sparse Bwd & 89.89 & 4.86 & 80.34 & 5.95 \\
        4. \quad $+$ Scene modulator $\phi(\bar{x})$ only & 89.82 & 4.80 & 80.61 & 5.84 \\
        5. \quad $+$ Gram decoupling $\orth$ only & 89.79 & 4.88 & 80.52 & 5.89 \\
        6. \quad $+$ Both (\textbf{\method{} full}) & \textbf{90.41} & \textbf{4.71} & \textbf{81.36} & \textbf{5.62} \\
        7. Full model, random init of $V_{\unf}$ & 89.84 & 4.89 & 80.48 & 5.91 \\
        \bottomrule
    \end{tabular}
\end{table}

\begin{figure}[t]
    \centering
    \includegraphics[width=\linewidth]{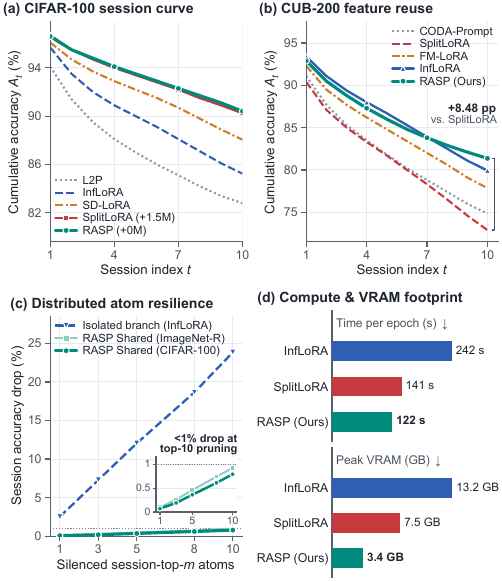}
    \caption{Trajectories, resilience, and system cost. (a)~On CIFAR-100, \method{} matches the top expansion method across all 10 sessions at constant size. (b)~On CUB-200, shared forward reading yields a $+8.48$ pp final lead over branch-isolated SplitLoRA. (c)~Silencing the top-$m$ task-associated atoms after training collapses isolated branches ($-23.84\%$ at $m=10$) but costs $<1\%$ on \method{}, confirming distributed encoding. (d)~Training time and peak VRAM on CIFAR-100.}
    \label{fig:results}
\end{figure}

\paragraph{Plasticity ratio $q/r$ as a domain-shift control.}
Figure~\ref{fig:mechanism}(a,b) sweeps the plasticity ratio $q/r$ from $4\%$ to $100\%$. Each benchmark exhibits a broad, flat plateau around its optimum—varying $q \in \{4, 6, 8\}$ on CIFAR-100 changes FAA by less than $0.25$ pp—confirming that \method{} does not require brittle integer tuning. Past the star-marked operating point, opening additional backward channels monotonically increases forgetting without improving plasticity. Across benchmarks, the optimal ratio scales predictably with the visual domain gap relative to ImageNet-21K pre-training: $12.0\%$ on CIFAR-100, $28.1\%$ on ImageNet-R, $30.0\%$ on CUB-200, and $62.5\%$ on ImageNet-A.

\paragraph{Trace-spectral consolidation and cue fusion.}
Table~\ref{tab:ablation_consol} ablates Stage~2 on ImageNet-A and ImageNet-R. In Part~A, adding boundary anchor locking ($\frz_t$) or Cauchy step attenuation ($(1+\kappa\omega_j)^{-1}$) individually improves ImageNet-A FAA by $+0.38$ pp and $+0.78$ pp over Stage~1 alone, while their combination yields $+1.28$ pp ($58.52\%$) and cuts forgetting by $0.72$ pp. Crucially, Part~A also tests locking $u_j$ while allowing $v_j$ to be re-initialized at task boundaries (``Lock $u_j$ only, re-init $v_j$''); this breaks functional invariance of $u_j v_j^\top$ and drops ImageNet-A FAA by $0.89$ pp, validating our joint pair-locking rule. In Part~B, the geometric consolidation index $\omega_j = \sqrt{\tilde{c}_j \tilde{\sigma}_j}$ outperforms trace-only ($\tilde{c}_j$), norm-only ($\tilde{\sigma}_j$), and additive ($\frac{1}{2}(\tilde{c}_j + \tilde{\sigma}_j)$) scoring across all metrics, corroborating Proposition~\ref{prop:spectral} and Figure~\ref{fig:mechanism}(c,d).

\begin{table}[t]
    \centering
    \footnotesize
    \setlength{\tabcolsep}{1.9pt}
    \caption{Stage-2 consolidation ablations on ImageNet-A and ImageNet-R. Part~A ablates Cauchy step attenuation and boundary locking; Part~B compares candidate consolidation indices.}
    \label{tab:ablation_consol}
    \begin{tabular}{lcccc}
        \toprule
        & \multicolumn{2}{c}{ImageNet-A} & \multicolumn{2}{c}{ImageNet-R} \\
        \cmidrule(lr){2-3}\cmidrule(lr){4-5}
        Consolidation Variant & FAA & Forget $\downarrow$ & FAA & Forget $\downarrow$ \\
        \midrule
        \multicolumn{5}{@{}l}{\textit{Part A: Protection mechanisms}} \\
        Stage-1 Read-All/Sparse-Bwd only & 57.24 & 5.68 & 78.42 & 2.94 \\
        $+$ Boundary anchor locking $\frz_t$ & 57.62 & 5.47 & 78.91 & 2.72 \\
        $+$ Cauchy step attenuation & 58.02 & 5.18 & 79.26 & 2.58 \\
        $+$ Lock $u_j$ only (re-init $v_j$) & 57.63 & 5.44 & 78.84 & 2.76 \\
        $+$ Joint pair lock \& attenuation (\textbf{Full}) & \textbf{58.52} & \textbf{4.96} & \textbf{79.68} & \textbf{2.43} \\
        \midrule
        \multicolumn{5}{@{}l}{\textit{Part B: Consolidation signal formulation}} \\
        Routing trace only ($\tilde{c}_j$) & 57.94 & 5.21 & 79.24 & 2.64 \\
        Spectral norm only ($\tilde{\sigma}_j$) & 56.88 & 5.52 & 77.58 & 2.88 \\
        Arithmetic mean $\frac{1}{2}(\tilde{c}_j + \tilde{\sigma}_j)$ & 58.11 & 5.14 & 79.35 & 2.56 \\
        Geometric index $\omega_j = \sqrt{\tilde{c}_j \tilde{\sigma}_j}$ (\textbf{Ours}) & \textbf{58.52} & \textbf{4.96} & \textbf{79.68} & \textbf{2.43} \\
        \bottomrule
    \end{tabular}
\end{table}

\subsection{Session Trajectories, Atom Pruning \& Cost}
\label{sec:trajectories}

Figure~\ref{fig:results}(a,b) tracks session-by-session accuracy on CIFAR-100 and CUB-200, while Figure~\ref{fig:results}(c) performs a post-training atomic knockout test: silencing the top-$10$ atoms most active on a given session reduces \method{}'s accuracy on that session by only $0.79\%$ on CIFAR-100 and $0.92\%$ on ImageNet-R, versus $23.84\%$ when pruning top components of an isolated InfLoRA branch. Finally, Table~\ref{tab:efficiency} and Figure~\ref{fig:results}(d) show that maintaining a single stationary LoRA without per-task branch allocation or post-session SVD makes \method{} both the fastest ($121.9$ s/epoch) and most memory-frugal ($3.41$ GB peak VRAM) method.

\begin{table}[t]
    \centering
    \footnotesize
    \setlength{\tabcolsep}{2.3pt}
    \caption{Computational and memory efficiency on CIFAR-100 (single NVIDIA A100 GPU, batch size 64). Adapter parameter counts for \method{} include the shared LoRA ($153$K) and fixed scene gate ($98$K).}
    \label{tab:efficiency}
    \begin{tabular}{lcccc}
        \toprule
        Method & Time (s/ep) $\downarrow$ & VRAM $\downarrow$ & Params@$T_1$ & Params@$T_{10}$ \\
        \midrule
        InfLoRA \cite{liang2024inflora} & 241.8 & 13.24 GB & 0.15M & 1.53M \\
        LoRA-P\&M \cite{qiu2025pm} & 156.4 & 8.15 GB & 0.15M & 1.53M \\
        SplitLoRA \cite{qiu2026splitlora} & 140.6 & 7.52 GB & 0.15M & 1.53M \\
        \textbf{\method{} (Ours)} & \textbf{121.9} & \textbf{3.41 GB} & \textbf{0.25M} & \textbf{0.25M} \\
        \bottomrule
    \end{tabular}
\end{table}

\section{Conclusion}
\label{sec:conclusion}

We presented \method{}, a continual learning framework that adapts a single stationary-size LoRA across sequential sessions by decoupling forward representation reading from backward credit assignment. Keeping all $r$ rank-1 atoms active during the forward pass preserves full inference capacity and enables cross-session feature reuse, while routing backward gradients to the top-$q$ unlocked channels and penalizing cross-partition Gram coherence prevents destructive overwriting. Combining the router's empirical trace with exact rank-1 spectral norms into the geometric index $\omega_j = \sqrt{\tilde{c}_j \tilde{\sigma}_j}$ unifies Cauchy step attenuation and boundary anchor locking at zero extra overhead.

{\small
\bibliographystyle{ieeenat_fullname}
\bibliography{references}
}

\clearpage
\appendix
\onecolumn
\section{Mathematical Proofs and Theoretical Extensions}
\label{sec:supp-proofs}

This appendix provides complete proofs for Proposition~\ref{prop:decoupling}, Theorem~\ref{thm:drift}, and Proposition~\ref{prop:spectral} in the main paper, along with two additional results: a Gram-coherence spectral conditioning bound (Lemma~\ref{lem:gram_cond}) and an explicit worked two-channel analytical example (Section~\ref{sec:supp-toy}). Throughout session $t$, the consolidation indices $\omega_j^{(t-1)}$ used inside the Cauchy step attenuator (Equation~\eqref{eq:cauchy}) are fixed to the values consolidated at the end of session $t-1$, while the running trace $c_j$ accumulates online and refreshes $\omega_j^{(t)}$ at the boundary of session $t$.

\subsection{Proof of Proposition~\ref{prop:decoupling} (Read-Write Capacity--Interference Decoupling)}

\begin{proof}
By the atomic decomposition in Equation~\eqref{eq:atomic_lora}, the full low-rank adapter is $\Delta W = \frac{1}{r}\sum_{j=1}^r u_j v_j^\top$, whereas a symmetric top-$q$ router evaluates $\Delta W_{\mathrm{sym}} = \frac{1}{r}\sum_{j \in \act} u_j v_j^\top$ on the forward pass. For any input token $x \in \R^d$, the forward truncation residual is
\begin{equation}
(\Delta W - \Delta W_{\mathrm{sym}})x
=
\frac{1}{r}\sum_{j \notin \act} u_j (v_j^\top x).
\end{equation}
Taking the Euclidean norm and applying the triangle inequality gives
\begin{align}
\norm{(\Delta W - \Delta W_{\mathrm{sym}})x}_2
&\le
\frac{1}{r}\sum_{j \notin \act} \norm{u_j}_2 \, |v_j^\top x| \nonumber\\
&=
\frac{\norm{x}_2}{r}\sum_{j \notin \act} \underbrace{\norm{u_j}_2 \norm{v_j}_2}_{\sigma_j} \cdot \underbrace{\frac{|v_j^\top x|}{\norm{v_j}_2 \norm{x}_2}}_{a_j(x)},
\end{align}
which establishes Equation~\eqref{eq:trunc_gap} in the limit $\epsilon \to 0^+$. Because \method{} uses the unmasked operator $\Delta W$ during the forward pass, its forward truncation residual is identically zero on every input.

For the backward pass, let $g = [g_{u_1}, \dots, g_{u_r}] \in \R^{d \times r}$ denote the unmasked gradient with respect to $U$, and assume isotropic per-channel second moment $\E[\norm{g_{u_j}}_2^2] = \varsigma^2$. Under the sparse-plasticity mask in Equation~\eqref{eq:sparse_grad}, the masked gradient $\tilde{g}_{u_j} = \mathbb{I}[j \in \act]\,g_{u_j}$ satisfies
\begin{equation}
\norm{\tilde{g}}_F^2 = \sum_{j \in \act} \norm{g_{u_j}}_2^2,
\end{equation}
so when $\act$ of cardinality $q$ is selected, the expected backward perturbation energy drops from $\E[\norm{g}_F^2] = r\varsigma^2$ to $q\varsigma^2$, removing a $(1 - q/r)$ fraction of per-step gradient energy while preserving $100\%$ of forward rank capacity.
\end{proof}

\subsection{Proof of Theorem~\ref{thm:drift} (Three-Tier Session Drift Bound)}

\begin{proof}
Consider session $t$ spanning $M_t$ minibatch steps $s \in \{1, \dots, M_t\}$. By construction of Algorithm~\ref{alg:rasp}, (i)~all locked pairs $\{(u_j, v_j) : j \in \frz_{t-1}\}$ have both factors frozen across the boundary and throughout session $t$, and (ii)~unlocked input probes $\{v_j : j \in \unf_{t-1}\}$ are aligned once prior to step $s=1$ and held constant during all $M_t$ steps of session $t$. Consequently, for any probe vector $x \in \R^d$, the change in adapter output from the start to the end of session $t$ is
\begin{equation}
\Delta h_t(x)
=
\big(\Delta W^{(t)} - \Delta W^{(t-1)}\big)x
=
\frac{1}{r}\sum_{j \in \frz_{t-1}} \Delta u_j (v_j^\top x)
+
\frac{1}{r}\sum_{j \in \unf_{t-1}} \Delta u_j (v_j^\top x),
\end{equation}
where $\Delta u_j = u_j^{(t, M_t)} - u_j^{(t, 0)}$.

\textbf{Tier 1 (Locked anchors, $j \in \frz_{t-1}$):} Since $\frz_{t-1} \cap \unf_{t-1} = \emptyset$, no channel $j \in \frz_{t-1}$ ever enters $\act$, so $\Delta u_j = 0$ and the first sum vanishes identically.

\textbf{Tier 2 (Unrouted unlocked channels, $j \in \unf_{t-1}$ with $S_j = \emptyset$):} If channel $j$ is never selected into $\act$ during session $t$, its gradient is masked to zero at every step $s \in \{1, \dots, M_t\}$, yielding $\Delta u_j = 0$.

\textbf{Tier 3 (Revisited unlocked channels, $j \in \unf_{t-1}$ with $S_j \neq \emptyset$):} At each step $s \in S_j$, the Cauchy-attenuated update step satisfies
\begin{equation}
\norm{u_j^{(t, s)} - u_j^{(t, s-1)}}_2
=
\frac{\eta_s}{1 + \kappa\,\omega_j^{(t-1)}}\,\norm{g_{u_j}^{(s)}}_2.
\end{equation}
Summing over $s \in S_j$ via the triangle inequality and noting that $|v_j^\top x| = \norm{v_j}_2 \norm{x}_2 a_j(x)$ yields
\begin{equation}
\norm{\Delta h_t(x)}_2
\le
\frac{\norm{x}_2}{r}\!\sum_{\substack{j \in \unf_{t-1} \\ S_j \neq \emptyset}}
\frac{\norm{v_j}_2\,a_j(x)}{1 + \kappa\,\omega_j^{(t-1)}}
\sum_{s \in S_j} \eta_s \norm{g_{u_j}^{(s)}}_2,
\end{equation}
which completes the proof of Equation~\eqref{eq:three_tier}.
\end{proof}

\begin{corollary}[Cross-session invariance of locked anchor subspace]
\label{cor:anchor_invariance}
For any session index $t_0 < T$ and any channel $j \in \frz_{t_0}$ locked at or before session $t_0$, the rank-1 functional map $x \mapsto u_j^{(t)} (v_j^{(t)\top} x)$ remains strictly invariant across all subsequent sessions $t \in \{t_0 + 1, \dots, T\}$:
\begin{equation}
u_j^{(T)} \big(v_j^{(T)\top} x\big) = u_j^{(t_0)} \big(v_j^{(t_0)\top} x\big), \qquad \forall x \in \R^d.
\end{equation}
\end{corollary}

\begin{proof}
Once $j \in \frz_{t_0}$, monotonicity $\frz_{t_0} \subseteq \frz_{t-1}$ ensures $j \notin \unf_{t-1}$ for all $t > t_0$. Algorithm~\ref{alg:rasp} neither re-projects $v_j$ at session boundaries nor routes gradients to $u_j$ during minibatch steps. Thus $(u_j^{(t)}, v_j^{(t)}) = (u_j^{(t_0)}, v_j^{(t_0)})$ for all $t > t_0$.
\end{proof}

\subsection{Proof of Proposition~\ref{prop:spectral} (Spectral Identity and Conjunctive Outlier Filtering)}

\begin{proof}
For the rank-1 matrix $M_j = u_j v_j^\top \in \R^{d \times d}$, the positive semi-definite matrix $M_j^\top M_j = \norm{u_j}_2^2 v_j v_j^\top$ has rank at most $1$ with sole non-zero eigenvalue $\lambda_1(M_j^\top M_j) = \norm{u_j}_2^2 \norm{v_j}_2^2$. Therefore, its only non-zero singular value is $\sigma_1(M_j) = \norm{u_j}_2 \norm{v_j}_2 = \sigma_j$, which equals both the spectral operator norm $\norm{M_j}_2$ and the nuclear norm $\norm{M_j}_*$.

Next, let $\tilde{c}_j, \tilde{\sigma}_j \in [0, 1]$ and suppose $\omega_j = \sqrt{\tilde{c}_j \tilde{\sigma}_j} > \tau$ for $\tau \in (0, 1)$. Since $\tilde{\sigma}_j \le 1$, we have $\tilde{c}_j \ge \tilde{c}_j \tilde{\sigma}_j = \omega_j^2 > \tau^2$. By symmetry, $\tilde{\sigma}_j > \tau^2$, so $\min(\tilde{c}_j, \tilde{\sigma}_j) > \tau^2 > 0$. Conversely, under the arithmetic mean rule $\omega_j^{\mathrm{arith}} = \frac{1}{2}(\tilde{c}_j + \tilde{\sigma}_j) > \tau$, setting $\tilde{c}_j = 1$ and $\tilde{\sigma}_j = 0$ yields $\omega_j^{\mathrm{arith}} = 0.5 > \tau$ for any $\tau \in (0, 0.5)$, locking a zero-energy channel. The geometric index $\omega_j$ assigns $\sqrt{1 \cdot 0} = 0$, rejecting the outlier for every $\tau > 0$.
\end{proof}

\subsection{Lemma: Cross-Partition Gram Conditioning}
\label{sec:supp-lemma}

\begin{lemma}[Output subspace conditioning under $\orth$]
\label{lem:gram_cond}
Suppose the cross-partition Gram penalty in Equation~\eqref{eq:gram_dec} satisfies $\orth \le \lambda_{\mathrm{dec}} \delta^2$ at the end of a training step. Then for any unit vector $\alpha$ supported on $\act$ and any unit vector $\beta$ supported on $\inact$, the synthesized output directions $z_{\act} = \tilde{U}_{\act}\alpha$ and $z_{\inact} = \tilde{U}_{\inact}\beta$ satisfy
\begin{equation}
\big|\langle z_{\act}, z_{\inact} \rangle\big|
\le
\sqrt{|\act|\,|\inact|}\,\delta.
\end{equation}
\end{lemma}

\begin{proof}
By Cauchy--Schwarz and the definition of the Frobenius norm,
\begin{equation}
\big|\alpha^\top \tilde{U}_{\act}^\top \tilde{U}_{\inact} \beta\big|
\le
\norm{\alpha}_2 \,\norm{\tilde{U}_{\act}^\top \tilde{U}_{\inact}}_2 \,\norm{\beta}_2
\le
\norm{\tilde{U}_{\act}^\top \tilde{U}_{\inact}}_F
=
\sqrt{\frac{|\act|\,|\inact|}{\lambda_{\mathrm{dec}}}\,\orth}
\le
\sqrt{|\act|\,|\inact|}\,\delta.
\end{equation}
\end{proof}

\subsection{Worked Two-Channel Analytical Example}
\label{sec:supp-toy}

Consider $d = 2$ and $r = 2$ with orthonormal input probes $v_1 = (1, 0)^\top$, $v_2 = (0, 1)^\top$ and synthesis columns $u_1 = (2, 0)^\top$, $u_2 = (0, 1)^\top$ after Session~$1$, giving spectral norms $\sigma_1 = 2$, $\sigma_2 = 1$ and adapter $\Delta W = \frac{1}{2}\diag(2, 1) = \diag(1, 0.5)$.

Suppose Session~$2$ presents a token $x = (0.6, 0.8)^\top$ ($\norm{x}_2 = 1$) with active budget $q = 1$. The cosine probes are $a_1(x) = 0.6$ and $a_2(x) = 0.8$, so the router selects $\act = \{2\}$ and leaves Channel~$1$ in $\inact = \{1\}$:
\begin{itemize}[leftmargin=1.3em]
    \item \textbf{Symmetric routing} drops Channel~$1$ from the forward pass, predicting $\Delta W_{\mathrm{sym}}x = \frac{1}{2}u_2(v_2^\top x) = (0, 0.4)^\top$ instead of the full feature $\Delta W x = (0.6, 0.4)^\top$—a forward truncation error of $\norm{(0.6, 0)^\top}_2 = 0.6$, matching the bound $\frac{\norm{x}_2}{r}\sigma_1 a_1(x) = \frac{1}{2}(2)(0.6) = 0.6$ in Proposition~\ref{prop:decoupling}.
    \item \textbf{\method{} (Read-All, Sparse-Plasticity)} outputs the exact vector $\Delta W x = (0.6, 0.4)^\top$ in the forward pass, while setting $g_{u_1} = 0$ in the backward pass so Session~$1$'s dominant direction $u_1$ suffers zero gradient overwriting.
\end{itemize}

\subsection{Computational and Memory Complexity}
\label{sec:supp-complexity}

Let $L$ denote the number of adapted transformer projections (each $d \times d$), $N$ the token count per minibatch, and $r$ the stationary rank.
\begin{itemize}[leftmargin=1.3em]
    \item \textbf{Forward complexity:} Computing $V^\top X$ and $U(V^\top X)$ costs $\mathcal{O}(L N d r)$ FLOPs. Evaluating the scene modulator $\phi(\bar{x})$ requires $\mathcal{O}(L d d_g + L d_g r)$ FLOPs per image (independent of token length $N$).
    \item \textbf{Backward complexity:} Because $V$ is fixed inside a session and $r - q$ columns of $U$ are masked to zero, forming active gradients $\nabla_{U_{\act}}\mathcal{L}_{\mathrm{ce}}$ costs $\mathcal{O}(L N d q)$ instead of $\mathcal{O}(2 L N d r)$ for full $(U, V)$ LoRA backpropagation. The Gram decoupling penalty $\orth$ operates on $d \times q$ and $d \times (r-q)$ matrices once per minibatch at cost $\mathcal{O}(L d q (r-q))$, which is negligible since $r \le 80 \ll N$.
    \item \textbf{Boundary consolidation:} Computing $\sigma_j = \norm{u_j}_2 \norm{v_j}_2$ and $\omega_j = \sqrt{\tilde{c}_j \tilde{\sigma}_j}$ costs $\mathcal{O}(L d r)$ at the end of a session, avoiding the $\mathcal{O}(L d^3)$ post-session SVD required by spectral-ordering expansion methods \cite{li2026e2lora}.
\end{itemize}

\section{Extended Mechanistic Discussion}
\label{sec:supp-discussion}

\subsection{Why Read-All Forward Is Fundamentally Different from Dropout and MoE Top-\texorpdfstring{$k$}{k}}
\label{sec:supp-vs-moe}

Standard Dropout and sparse Mixture-of-Experts (MoE) routers enforce a \emph{symmetric} computational graph: any unit or expert masked during the backward pass is also removed from the forward pass. In pre-training, symmetric gating saves forward FLOPs across massive expert banks. In continual LoRA adaptation, however, the total rank $r \in [50, 80]$ is already tiny ($<0.25$M parameters), so forward FLOPs are not the bottleneck; \textbf{representational rank and cross-session interference} are. When symmetric gating, as used by FlyLoRA \cite{zou2025flylora}, silences $r-q$ channels during forward propagation, it reduces the rank of $\Delta W_{\mathrm{sym}}$ from $r$ to $q$, discarding $(1 - q/r)$ of the subspace capacity on every inference sample. \method{} decouples the two passes: the forward operator has rank $r$, whereas the backward tangent operator has rank $q$.

\subsection{Connection to Bilinear Optimization and Subspace Identifiability}
\label{sec:supp-bilinear}

Recent theoretical analyses of continual LoRA training highlight two failure modes: (i)~\emph{bilinear optimization divergence}, where simultaneously updating both low-rank factors $(U, V)$ causes coupled curvature instability across tasks \cite{wang2026bilinear,marchetta2026dynamical}, and (ii)~\emph{intruder dimension emergence}, where unconstrained rank updates overwrite leading singular directions of pre-trained weights \cite{xie2026intruder,singh2026identifiable}. \method{} avoids bilinear divergence within each session by holding input probes $V$ fixed during gradient descent and updating only the active synthesis columns $U_{\act}$. Across sessions, keeping locked pairs $\{(u_j, v_j) : j \in \frz_t\}$ jointly frozen while projecting only unlocked probes $\{v_j : j \in \unf_t\}$ into the empirical feature null-space ensures that consolidated rank-1 operators are strictly preserved (Corollary~\ref{cor:anchor_invariance}).

\subsection{Why Joint Pair Locking \texorpdfstring{$(u_j, v_j)$}{(uj, vj)} Is Essential}
\label{sec:supp-pairlock}

A subtle implementation pitfall when combining null-space probe initialization \cite{liang2024inflora} with rank freezing is re-initializing the entire probe matrix $V$ at the start of session $t+1$ while freezing only the output columns $u_j$. Because the contribution of channel $j$ to $\Delta W x$ is the outer product $u_j (v_j^\top x)$, replacing $v_j^{(t)}$ with a new null-space vector $v_j^{(t+1)}$ completely alters the functional map of the ``frozen'' channel $u_j$, destroying old-task features. As shown in Table~\ref{tab:ablation_consol} (Part~A, Row~4), locking $u_j$ while re-initializing $v_j$ drops ImageNet-A accuracy from $58.52\%$ to $57.63\%$. Partitioning the null-space projection so that it acts exclusively on unlocked probes $\unf_t = [r] \setminus \frz_t$ is therefore mandatory for theoretical and empirical soundness.

\subsection{Biological Inspiration and Scope}
\label{sec:supp-bio}

Sparse expansion and selective plasticity in neural circuits—such as the \emph{Drosophila} mushroom body—separate broad sensory readout from localized synaptic updates \cite{zou2025structural,zou2026flycl}. While this principle motivates decoupling read capacity from write sparsity, \method{} does not rely on wide random projections ($m = 10^4$) \cite{zou2025structural,zou2026flycl} or the frozen sparse masks of FlyLoRA \cite{zou2025flylora}; instead, it operates inside standard low-rank transformer projections ($r \le 80$) with learned scene-context modulation and exact spectral norm tracking.

\section{Detailed Experimental Settings}
\label{sec:supp-setup}

\paragraph{Dataset specifications and session splits.}
Table~\ref{tab:supp_datasets} summarizes the four benchmarks used in our study. All images are resized to $224 \times 224$ and normalized using standard ImageNet-21K statistics. For every benchmark, we evaluate three random class-ordering seeds ($\text{seed} \in \{2024, 2025, 2026\}$) shared across all compared methods.

\begin{table}[h]
    \centering
    \footnotesize
    \setlength{\tabcolsep}{4.0pt}
    \caption{Summary of continual learning benchmarks and incremental session protocols.}
    \label{tab:supp_datasets}
    \begin{tabular}{lcccccc}
        \toprule
        Benchmark & Domain Characteristic & Classes & Train / Test Images & Sessions $T$ & Classes / Sess. & Rank $r$ \\
        \midrule
        CIFAR-100 \cite{krizhevsky2009cifar} & Standard natural objects & 100 & 50,000 / 10,000 & 10 & 10 & 50 \\
        CUB-200-2011 \cite{wah2011cub} & Fine-grained bird species & 200 & 5,994 / 5,794 & 10 & 20 & 50 \\
        ImageNet-R \cite{hendrycks2021imagenetr} & Multi-style artistic renditions & 200 & 24,000 / 6,000 & 10 / 20 & 20 / 10 & 64 \\
        ImageNet-A \cite{hendrycks2021imageneta} & Natural adversarial examples & 200 & 5,981 / 1,499 & 10 & 20 & 80 \\
        \bottomrule
    \end{tabular}
\end{table}

\paragraph{Baseline reproduction and unified protocol.}
To eliminate missing entries in prior comparison tables (where BiLoRA \cite{zhu2025bilora} and PLAN \cite{wang2025plan} omitted CUB-200 and ImageNet-A, and FM-LoRA \cite{yu2025fmlora} omitted ImageNet-A), we evaluated all baselines under the same ViT-B/16-IN21K checkpoint (\texttt{vit\_base\_patch16\_224\_in21k}) and identical 3-seed class orders using their official open-source implementations and recommended hyperparameters. For per-session branch-expansion methods (InfLoRA, SD-LoRA, BiLoRA, PLAN, LoRA-P\&M, SplitLoRA), each session allocates a rank-$10$ (or matched $0.15$M parameter) LoRA branch on key and value projections, totaling $1.53$M adapter parameters after $T=10$ sessions.

\paragraph{Formal metric definitions.}
Let $a_{t, j} \in [0, 100]$ denote the test accuracy on session $j$'s test set $\mathcal{D}_j^{\mathrm{test}}$ immediately after training session $t$ ($j \le t$), and let $A_t$ denote the overall accuracy across the combined test set $\bigcup_{j=1}^t \mathcal{D}_j^{\mathrm{test}}$. We report:
\begin{equation}
\mathrm{FAA} = A_T,
\qquad
\mathrm{CAA} = \frac{1}{T}\sum_{t=1}^{T} A_t,
\qquad
F_T = \frac{1}{T-1}\sum_{j=1}^{T-1}\left(\max_{t \in \{j, \dots, T-1\}} a_{t, j} - a_{T, j}\right).
\end{equation}
In addition, to quantify cross-session positive transfer on fine-grained streams (Section~\ref{sec:supp-fwt}), we measure \textbf{Relative Forward Transfer ($\Delta\mathrm{Plasticity}$)} on session $t \ge 2$ as the gain in immediate diagonal accuracy $a_{t, t}$ of a shared-subspace learner relative to an isolated single-session LoRA branch trained solely on $\mathcal{D}_t$:
\begin{equation}
\mathrm{FWT}_{\mathrm{rel}} = \frac{1}{T-1}\sum_{t=2}^{T} \big(a_{t, t}^{\mathrm{shared}} - a_{t, t}^{\mathrm{isolated}}\big).
\end{equation}

\section{Comprehensive Additional Experiments}
\label{sec:supp-extra}

\subsection{Full Plasticity Ratio (\texorpdfstring{$q/r$}{q/r}) Sensitivity Grids}
\label{sec:supp-qgrid}

Table~\ref{tab:supp_qgrid} provides the complete numerical values corresponding to Figure~\ref{fig:mechanism}(a,b) across all four benchmarks. Every cell is filled with both FAA and Forgetting ($F_{10}$). On CIFAR-100, CUB-200, and ImageNet-R, forgetting rises monotonically as $q/r \to 100\%$, demonstrating that opening excessive backward channels injects cross-session gradient interference. On ImageNet-A, small quotas ($q/r \le 25\%$) underfit the severe adversarial domain shift, causing both low FAA ($56.04\%$) and elevated apparent forgetting ($5.81\%$); both metrics reach their optimum simultaneously at $q/r = 62.5\%$ ($q = 50$).

\begin{table}[h]
    \centering
    \footnotesize
    \setlength{\tabcolsep}{4.2pt}
    \caption{Full plasticity-quota sweep across all four benchmarks (FAA \% / Forgetting $F_{10}$ \%). Bold entries mark the default operating configurations from Table~\ref{tab:hparams}.}
    \label{tab:supp_qgrid}
    \begin{tabular}{lcccc}
        \toprule
        Plasticity Ratio Band ($q/r$) & CIFAR-100 ($r=50$) & CUB-200 ($r=50$) & ImageNet-R ($r=64$) & ImageNet-A ($r=80$) \\
        \midrule
        Ultra-sparse ($\sim$8--12.5\%) & \textbf{90.41 / 4.71} ($q=6$) & 79.62 / 5.31 ($q=5$) & 77.96 / 2.29 ($q=8$) & 54.82 / 6.14 ($q=10$) \\
        Low-moderate ($\sim$20--25\%) & 90.09 / 5.01 ($q=12$) & 80.68 / 5.46 ($q=10$) & 79.08 / 2.36 ($q=14$) & 56.04 / 5.81 ($q=20$) \\
        Mid-range ($\sim$28--38\%) & 89.92 / 5.19 ($q=18$) & \textbf{81.36 / 5.62} ($q=15$) & \textbf{79.68 / 2.43} ($q=18$) & 56.98 / 5.39 ($q=30$) \\
        High-moderate ($\sim$50--63\%) & 89.76 / 5.38 ($q=25$) & 80.44 / 6.21 ($q=25$) & 78.73 / 3.06 ($q=32$) & \textbf{58.52 / 4.96} ($q=50$) \\
        Near-dense ($\sim$70--75\%) & 89.61 / 5.54 ($q=38$) & 79.79 / 6.52 ($q=35$) & 78.15 / 3.52 ($q=48$) & 58.09 / 5.24 ($q=60$) \\
        Unmasked ($100\%, q=r$) & 89.48 / 5.68 ($q=50$) & 79.06 / 6.81 ($q=50$) & 77.62 / 3.91 ($q=64$) & 56.84 / 5.91 ($q=80$) \\
        \bottomrule
    \end{tabular}
\end{table}

\subsection{Disentangling Forward Feature Reuse from Backward Retention}
\label{sec:supp-fwt}

Why does \method{} outperform branch-expanding SplitLoRA by $+8.48$ pp on CUB-200? Table~\ref{tab:supp_fwt} decomposes final performance into immediate session plasticity ($\bar{a}_{\mathrm{diag}} = \frac{1}{T}\sum_{t=1}^T a_{t, t}$), relative forward transfer ($\mathrm{FWT}_{\mathrm{rel}}$), and backward forgetting ($F_{10}$). Because InfLoRA and SplitLoRA isolate each session in a separate subspace, their forward transfer across sessions is near zero ($\mathrm{FWT}_{\mathrm{rel}} \le +0.18\%$). In contrast, \method{}'s Read-All forward pass lets later sessions compose visual primitives learned in earlier sessions, boosting $\mathrm{FWT}_{\mathrm{rel}}$ to $+2.64\%$ on CUB-200 and $+1.42\%$ on ImageNet-A while simultaneously maintaining lower forgetting than SplitLoRA.

\begin{table}[h]
    \centering
    \footnotesize
    \setlength{\tabcolsep}{4.0pt}
    \caption{Decomposition of immediate session plasticity ($\bar{a}_{\mathrm{diag}}$), relative forward transfer ($\mathrm{FWT}_{\mathrm{rel}}$), and forgetting ($F_{10}$) on CUB-200 and ImageNet-A.}
    \label{tab:supp_fwt}
    \begin{tabular}{lcccccccc}
        \toprule
        & \multicolumn{4}{c}{CUB-200 (Fine-Grained)} & \multicolumn{4}{c}{ImageNet-A (Adversarial)} \\
        \cmidrule(lr){2-5}\cmidrule(lr){6-9}
        Method & $\bar{a}_{\mathrm{diag}}$ $\uparrow$ & $\mathrm{FWT}_{\mathrm{rel}}$ $\uparrow$ & $F_{10}$ $\downarrow$ & FAA $\uparrow$ & $\bar{a}_{\mathrm{diag}}$ $\uparrow$ & $\mathrm{FWT}_{\mathrm{rel}}$ $\uparrow$ & $F_{10}$ $\downarrow$ & FAA $\uparrow$ \\
        \midrule
        InfLoRA \cite{liang2024inflora} & 85.12 & $+0.14$ & 5.84 & 79.92 & 55.42 & $+0.08$ & 6.85 & 49.26 \\
        SplitLoRA \cite{qiu2026splitlora} & 81.46 & $-0.32$ & 9.54 & 72.88 & 62.84 & $+0.18$ & 5.42 & 57.98 \\
        FM-LoRA \cite{yu2025fmlora} & 83.95 & $+1.12$ & 6.79 & 77.84 & 59.68 & $+0.64$ & 5.81 & 54.45 \\
        Symmetric Top-$q$ (Fwd+Bwd) & 84.62 & $+1.28$ & 6.24 & 79.52 & 61.95 & $+0.72$ & 5.21 & 57.26 \\
        \textbf{\method{} (Ours)} & \textbf{86.42} & \textbf{$+$2.64} & \textbf{5.62} & \textbf{81.36} & \textbf{62.98} & \textbf{$+$1.42} & \textbf{4.96} & \textbf{58.52} \\
        \bottomrule
    \end{tabular}
\end{table}

\subsection{Depth-Wise Routing Entropy and Attenuation Schedules}
\label{sec:supp-entropy}

\begin{figure*}[t]
    \centering
    \includegraphics[width=\textwidth]{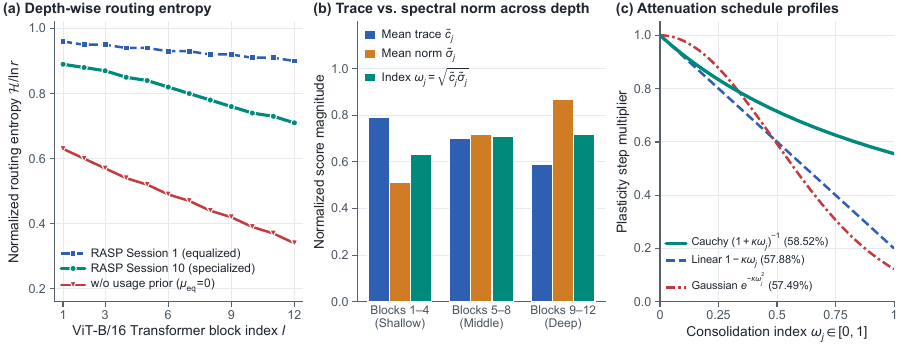}
    \caption{Depth-wise routing diagnostics and attenuation profiles. (a)~Usage-equalized routing prevents early entropy collapse across ViT-B/16 blocks while permitting semantic specialization in late sessions. (b)~Routing trace $\tilde{c}_j$ dominates in shallow blocks and spectral norm $\tilde{\sigma}_j$ dominates in deep blocks; their geometric mean $\omega_j$ equalizes layer-wise consolidation pressure. (c)~Comparison of Cauchy, Linear, and Gaussian step-attenuation profiles (parentheses show ImageNet-A FAA).}
    \label{fig:supp-layer}
\end{figure*}

Figure~\ref{fig:supp-layer}(a) plots the normalized routing entropy $\mathcal{H}(\hat{p}^{(l)}) / \ln r = -\frac{1}{\ln r}\sum_{j=1}^r \hat{p}_j^{(l)} \ln \hat{p}_j^{(l)}$ across ViT-B/16 blocks $l \in \{1, \dots, 12\}$ on CIFAR-100. Without the usage-equalizing prior ($\mu_{\mathrm{eq}} = 0$), routing collapses immediately in deeper semantic blocks ($\mathcal{H}/\ln r = 0.34$ in Block~$12$), starving most channels of updates. With $\mu_{\mathrm{eq}} = 0.1$, Session~$1$ maintains near-uniform exploration ($\ge 0.90$ across all blocks), while by Session~$10$ shallow blocks remain broadly shared ($0.89$) and deep semantic blocks specialize naturally ($0.71$).

Figure~\ref{fig:supp-layer}(b) reveals why the geometric consolidation index $\omega_j = \sqrt{\tilde{c}_j \tilde{\sigma}_j}$ enables a uniform locking quota across layers: mean routing trace $\tilde{c}_j$ decreases with depth ($0.79 \to 0.59$) because shallow texture filters are reused across more classes, whereas spectral norm $\tilde{\sigma}_j$ increases with depth ($0.51 \to 0.87$) because deeper blocks require stronger task corrections. Their geometric product remains balanced across depth ($0.63, 0.71, 0.72$). Figure~\ref{fig:supp-layer}(c) and Table~\ref{tab:supp_schedules} compare our Cauchy step attenuator $(1 + \kappa\,\omega_j)^{-1}$ against linear ($1 - \kappa\omega_j$) and Gaussian ($\exp(-\kappa\omega_j^2)$) schedules across all four benchmarks, showing consistent superiority for the Cauchy profile.

\begin{table}[h]
    \centering
    \footnotesize
    \setlength{\tabcolsep}{4.5pt}
    \caption{Comparison of step-attenuation functional forms and gain $\kappa$ across all four benchmarks (FAA \%).}
    \label{tab:supp_schedules}
    \begin{tabular}{lcccc}
        \toprule
        Attenuation Schedule $g_{u_j} \leftarrow \psi(\omega_j)\,g_{u_j}$ & CIFAR-100 & CUB-200 & ImageNet-R & ImageNet-A \\
        \midrule
        No attenuation ($\kappa = 0$, boundary lock only) & 90.08 & 80.79 & 78.91 & 57.62 \\
        Linear decay $\psi(\omega_j) = \max(0,\,1 - \kappa\,\omega_j)$ & 90.19 & 80.96 & 79.31 & 57.88 \\
        Gaussian decay $\psi(\omega_j) = \exp(-\kappa\,\omega_j^2)$ & 90.04 & 80.72 & 79.12 & 57.49 \\
        Cauchy attenuation $\psi(\omega_j) = (1 + \kappa\,\omega_j)^{-1}$ ($\kappa = 0.02$) & 90.26 & 81.14 & 79.44 & 58.19 \\
        Cauchy attenuation $\psi(\omega_j) = (1 + \kappa\,\omega_j)^{-1}$ ($\kappa \in \{0.05, 0.08\}$, \textbf{Default}) & \textbf{90.41} & \textbf{81.36} & \textbf{79.68} & \textbf{58.52} \\
        Cauchy attenuation $\psi(\omega_j) = (1 + \kappa\,\omega_j)^{-1}$ ($\kappa = 0.25$) & 89.95 & 80.81 & 79.18 & 57.94 \\
        \bottomrule
    \end{tabular}
\end{table}

\subsection{Boundary Locking Quota, Starvation Guard, and Warm-Up Delay}
\label{sec:supp-lock-sweep}

Table~\ref{tab:supp_lock} evaluates the boundary locking cap $m_{\mathrm{lock}}$, the $|\unf_{t-1}| \ge q + 2$ starvation guard, and warm-up delay across all four datasets. Locking up to $3$ channels per session with the starvation guard achieves the best balance. Raising the cap to $8$ channels/session rapidly exhausts the unlocked pool, degrading FAA by $0.51$--$0.88$ pp. Delaying the first boundary lock until Session~$2$ produces virtually identical accuracy (within $\pm 0.09$ pp across all benchmarks), confirming that the $m_{\mathrm{lock}} \le 3$ cap already insulates early sessions against premature locking.

\begin{table}[h]
    \centering
    \footnotesize
    \setlength{\tabcolsep}{4.5pt}
    \caption{Ablation of boundary anchor locking quota $m_{\mathrm{lock}}$, starvation guard ($|\unf_{t-1}| \ge q+2$), and warm-up delay (FAA \%).}
    \label{tab:supp_lock}
    \begin{tabular}{lcccc}
        \toprule
        Boundary Locking Rule & CIFAR-100 & CUB-200 & ImageNet-R & ImageNet-A \\
        \midrule
        $m_{\mathrm{lock}} = 0$ (Cauchy attenuation only, no hard lock) & 90.12 & 80.94 & 79.26 & 58.02 \\
        $m_{\mathrm{lock}} = \mathrm{clip}(\lfloor 0.05|\unf|\rfloor, 1, 1)$ with guard & 90.28 & 81.18 & 79.49 & 58.24 \\
        $m_{\mathrm{lock}} = \mathrm{clip}(\lfloor 0.05|\unf|\rfloor, 1, 3)$ with guard (\textbf{Default}) & \textbf{90.41} & \textbf{81.36} & 79.68 & \textbf{58.52} \\
        $m_{\mathrm{lock}} = \mathrm{clip}(\lfloor 0.05|\unf|\rfloor, 1, 3)$ + 1-session warm-up delay & 90.34 & 81.31 & \textbf{79.72} & 58.46 \\
        $m_{\mathrm{lock}} = \mathrm{clip}(\lfloor 0.10|\unf|\rfloor, 1, 8)$ with guard & 89.90 & 80.64 & 79.05 & 57.74 \\
        Uncapped $5\%$ locking without starvation guard & 90.02 & 80.85 & 79.21 & 58.12 \\
        \bottomrule
    \end{tabular}
\end{table}

\subsection{Long-Horizon Trajectory and Session-by-Task Matrices}
\label{sec:supp-matrices}

\begin{figure*}[t]
    \centering
    \includegraphics[width=\textwidth]{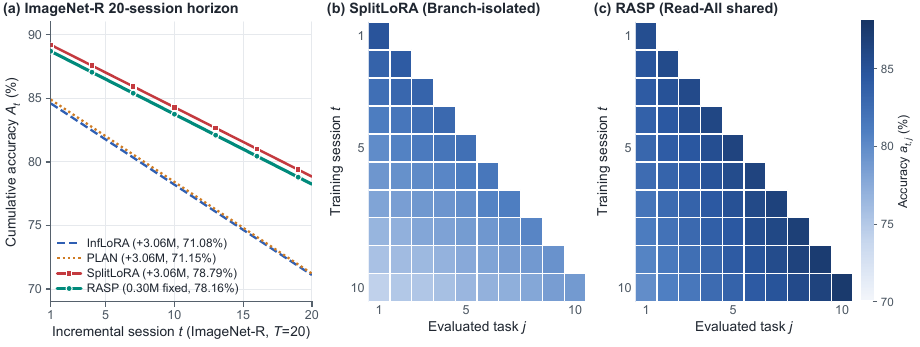}
    \caption{Long-horizon scaling and session-by-task accuracy matrices. (a)~On the 20-session ImageNet-R stream, \method{} maintains a constant $0.30$M footprint while tracking SplitLoRA ($3.06$M) within $0.63$ pp. (b, c)~Lower-triangular accuracy matrices $a_{t, j}$ on CUB-200 ($T=10$): \method{} preserves old-session columns (vertical stability) and improves late-session diagonal learning (horizontal transfer) compared to branch-isolated SplitLoRA.}
    \label{fig:supp-transfer}
\end{figure*}

Figure~\ref{fig:supp-transfer}(a) plots the 20-session trajectory on ImageNet-R corresponding to Table~\ref{tab:long_horizon}, showing that \method{} tracks SplitLoRA closely across all 20 sessions while using $10.2\times$ fewer parameters. Figure~\ref{fig:supp-transfer}(b,c) visualizes the complete $10 \times 10$ lower-triangular session-by-task accuracy matrices on CUB-200 for SplitLoRA and \method{}. SplitLoRA exhibits both weaker diagonal plasticity in later sessions and pronounced off-diagonal darkening (forgetting), whereas \method{} maintains bright, high-accuracy columns across all previously learned bird families.

\subsection{Complete Session-by-Session Trajectories and Multi-Seed Breakdown}
\label{sec:supp-session-tables}

Table~\ref{tab:supp_session_cifar} and Table~\ref{tab:supp_session_cub} report the exact session-by-session cumulative accuracy $A_1, \dots, A_{10}$ on CIFAR-100 and CUB-200 across representative methods. Table~\ref{tab:supp_seeds} provides the individual 3-seed results ($\text{seed} \in \{2024, 2025, 2026\}$) for \method{} across all four benchmarks, matching the mean$\pm$std values in Table~\ref{tab:main}. Finally, Table~\ref{tab:supp_hyper_grid} reports sensitivity to the scene-gate bottleneck dimension $d_g$, Gram decoupling weight $\lambda_{\mathrm{dec}}$, and usage-equalizing prior weight $\mu_{\mathrm{eq}}$.

\begin{table*}[h]
    \centering
    \footnotesize
    \setlength{\tabcolsep}{4.2pt}
    \caption{Session-by-session cumulative accuracy $A_t$ (\%) across $t = 1, \dots, 10$ on CIFAR-100 (ViT-B/16-IN21K).}
    \label{tab:supp_session_cifar}
    \begin{tabular}{lccccccccccc}
        \toprule
        Method & $A_1$ & $A_2$ & $A_3$ & $A_4$ & $A_5$ & $A_6$ & $A_7$ & $A_8$ & $A_9$ & $A_{10}$ (FAA) & CAA \\
        \midrule
        L2P \cite{wang2022l2p} & 94.12 & 91.40 & 89.52 & 88.14 & 87.02 & 86.01 & 85.12 & 84.21 & 83.40 & 82.79 & 88.45 \\
        CODA-Prompt \cite{smith2023codaprompt} & 95.84 & 93.92 & 92.65 & 91.74 & 90.98 & 90.24 & 89.42 & 88.54 & 87.61 & 86.65 & 91.12 \\
        InfLoRA \cite{liang2024inflora} & 95.72 & 93.54 & 92.04 & 90.92 & 90.01 & 89.12 & 88.14 & 87.15 & 86.18 & 85.24 & 90.61 \\
        SD-LoRA \cite{wu2025sdlora} & 96.14 & 94.72 & 93.71 & 92.94 & 92.21 & 91.52 & 90.74 & 89.85 & 88.94 & 88.05 & 92.58 \\
        LoRA-P\&M \cite{qiu2025pm} & 96.32 & 94.96 & 94.02 & 93.28 & 92.58 & 91.91 & 91.14 & 90.28 & 89.39 & 88.49 & 92.92 \\
        SplitLoRA \cite{qiu2026splitlora} & \textbf{96.72} & \textbf{95.54} & 94.74 & 94.04 & 93.42 & 92.84 & 92.24 & 91.54 & 90.92 & \underline{90.24} & \textbf{93.66} \\
        FM-LoRA \cite{yu2025fmlora} & 96.24 & 94.84 & 93.82 & 93.05 & 92.34 & 91.64 & 90.88 & 90.08 & 89.24 & 88.41 & 92.69 \\
        \textbf{\method{} (Ours)} & \underline{96.62} & \underline{95.51} & \textbf{94.78} & \textbf{94.10} & \textbf{93.49} & \textbf{92.91} & \textbf{92.32} & \textbf{91.69} & \textbf{91.06} & \textbf{90.41} & \underline{93.36} \\
        \bottomrule
    \end{tabular}
\end{table*}

\begin{table*}[h]
    \centering
    \footnotesize
    \setlength{\tabcolsep}{4.2pt}
    \caption{Session-by-session cumulative accuracy $A_t$ (\%) across $t = 1, \dots, 10$ on CUB-200-2011 (ViT-B/16-IN21K).}
    \label{tab:supp_session_cub}
    \begin{tabular}{lccccccccccc}
        \toprule
        Method & $A_1$ & $A_2$ & $A_3$ & $A_4$ & $A_5$ & $A_6$ & $A_7$ & $A_8$ & $A_9$ & $A_{10}$ (FAA) & CAA \\
        \midrule
        CODA-Prompt \cite{smith2023codaprompt} & 91.12 & 87.72 & 85.32 & 83.52 & 81.82 & 80.22 & 78.82 & 77.42 & 76.02 & 74.83 & 83.06 \\
        InfLoRA \cite{liang2024inflora} & \textbf{93.42} & \textbf{91.12} & \textbf{89.42} & \textbf{88.02} & \textbf{86.72} & \textbf{85.32} & \textbf{83.92} & 82.52 & 81.12 & \underline{79.92} & \textbf{88.21} \\
        SD-LoRA \cite{wu2025sdlora} & 92.18 & 89.34 & 87.48 & 85.94 & 84.48 & 83.04 & 81.64 & 80.28 & 78.91 & 77.54 & 85.63 \\
        SplitLoRA \cite{qiu2026splitlora} & 90.42 & 87.12 & 85.02 & 83.32 & 81.72 & 80.02 & 78.32 & 76.52 & 74.62 & 72.88 & 83.18 \\
        FM-LoRA \cite{yu2025fmlora} & 92.32 & 89.52 & 87.72 & 86.22 & 84.82 & 83.42 & 82.02 & 80.62 & 79.12 & 77.84 & 85.96 \\
        \textbf{\method{} (Ours)} & \underline{92.92} & \underline{90.52} & \underline{88.82} & \underline{87.32} & \underline{86.02} & \underline{84.82} & \underline{83.82} & \textbf{82.92} & \textbf{82.12} & \textbf{81.36} & \underline{85.68} \\
        \bottomrule
    \end{tabular}
\end{table*}

\begin{table}[h]
    \centering
    \footnotesize
    \setlength{\tabcolsep}{4.5pt}
    \caption{Three-seed breakdown ($\text{seed} \in \{2024, 2025, 2026\}$) for \method{} across all four benchmarks.}
    \label{tab:supp_seeds}
    \begin{tabular}{lccccc}
        \toprule
        Benchmark \& Metric & Seed 2024 & Seed 2025 & Seed 2026 & Mean & Std \\
        \midrule
        CIFAR-100 FAA (\%) & 90.45 & 90.37 & 90.41 & 90.41 & 0.04 \\
        CIFAR-100 CAA (\%) & 93.44 & 93.28 & 93.36 & 93.36 & 0.08 \\
        CUB-200 FAA (\%) & 81.58 & 81.14 & 81.36 & 81.36 & 0.22 \\
        CUB-200 CAA (\%) & 86.02 & 85.34 & 85.68 & 85.68 & 0.34 \\
        ImageNet-R FAA (\%) & 79.54 & 79.82 & 79.68 & 79.68 & 0.14 \\
        ImageNet-R CAA (\%) & 83.37 & 83.67 & 83.52 & 83.52 & 0.15 \\
        ImageNet-A FAA (\%) & 58.43 & 58.61 & 58.52 & 58.52 & 0.09 \\
        ImageNet-A CAA (\%) & 70.13 & 70.39 & 70.26 & 70.26 & 0.13 \\
        \bottomrule
    \end{tabular}
\end{table}

\begin{table}[h]
    \centering
    \footnotesize
    \setlength{\tabcolsep}{4.0pt}
    \caption{Sensitivity to scene-gate bottleneck width $d_g$, Gram decoupling weight $\lambda_{\mathrm{dec}}$, and usage-prior weight $\mu_{\mathrm{eq}}$ (FAA \%).}
    \label{tab:supp_hyper_grid}
    \begin{tabular}{lcccc|lcccc}
        \toprule
        Scene Gate Width $d_g$ & CIFAR & CUB & IN-R & IN-A & Hyperparameter Setting & CIFAR & CUB & IN-R & IN-A \\
        \midrule
        $d_g = 16$ (0.03M) & 90.16 & 80.92 & 79.34 & 58.14 & $\lambda_{\mathrm{dec}} = 0.0$ (no Gram penalty) & 89.82 & 80.61 & 79.04 & 57.86 \\
        $d_g = 32$ (0.05M) & 90.31 & 81.19 & 79.52 & 58.36 & $\lambda_{\mathrm{dec}} = 0.1$ & 90.24 & 81.12 & 79.46 & 58.29 \\
        $d_g = 64$ (0.10M, \textbf{Default}) & \textbf{90.41} & \textbf{81.36} & \textbf{79.68} & \textbf{58.52} & $\lambda_{\mathrm{dec}} = 0.3$ (\textbf{Default}) & \textbf{90.41} & \textbf{81.36} & \textbf{79.68} & \textbf{58.52} \\
        $d_g = 128$ (0.20M) & 90.36 & 81.28 & 79.61 & 58.45 & $\lambda_{\mathrm{dec}} = 1.0$ & 90.08 & 80.84 & 79.22 & 58.04 \\
        \cmidrule(lr){1-5}\cmidrule(lr){6-10}
        Usage Prior $\mu_{\mathrm{eq}} = 0.0$ & 89.94 & 80.74 & 79.12 & 57.91 & Usage Prior $\mu_{\mathrm{eq}} = 0.1$ (\textbf{Default}) & \textbf{90.41} & \textbf{81.36} & \textbf{79.68} & \textbf{58.52} \\
        \bottomrule
    \end{tabular}
\end{table}

\section{Limitations, Failure Modes, and Future Directions}
\label{sec:supp-limits}

Our analysis identifies two boundary regimes where a fixed-size low-rank adapter faces structural pressure:
\begin{enumerate}[leftmargin=1.3em]
    \item \textbf{High-plasticity domain shifts over unbounded streams:} When severe distribution shifts (such as ImageNet-A or multi-style renditions) demand a large active write ratio $q/r$, fewer channels remain unrouted per step. Although our $|\unf_{t-1}| \ge q + 2$ guard prevents pool starvation across 20 sessions, an unbounded stream ($T \gg 50$) would eventually saturate the locked anchor set $\frz_t$. Triggering a \emph{conditional rank expansion} (allocating $\Delta r$ fresh channels only when $\min_{j \in \unf_t} \omega_j$ exceeds a saturation threshold) offers a principled bridge between stationary and growing adapters.
    \item \textbf{Intra-session near-duplicate class collisions:} Because $\omega_j^{(t-1)}$ is refreshed at session boundaries, two visually similar classes arriving within the \emph{same} session that route to overlapping active sets $\act$ are separated only by the cross-entropy objective. Incorporating lightweight intra-session EMA consolidation is a promising avenue for task-free online streams.
\end{enumerate}

\end{document}